\documentclass[sigconf]{acmart}
\usepackage{gensymb}
\usepackage{soul,color}
\usepackage{booktabs}
\usepackage{multicol}
\usepackage{multirow}
\usepackage{longtable}
\usepackage{dsfont}
\usepackage{subcaption}
\usepackage{comment}
\usepackage{enumitem}

\AtBeginDocument{%
  }

\setcopyright{none}
\renewcommand\footnotetextcopyrightpermission[1]{}

\begin{document}

\title{Beyond Benchmarks: Continuous Edge Inference for Traffic Sign Recognition on Embedded Vehicular Hardware}

\author{Aditya Mishra}
\affiliation{%
  \institution{Indian Institute of Science Education and Research}
  \city{Bhopal}
  \country{India}}
\email{aditya21@iiserb.ac.in}

\author{Anushka Tailor}
\affiliation{%
  \institution{Indian Institute of Science Education and Research}
  \city{Bhopal}
  \country{India}}
\email{anushkat23@iiserb.ac.in}

\author{Haroon Lone}
\affiliation{%
  \institution{Indian Institute of Science Education and Research}
  \city{Bhopal}
  \country{India}}
  \email{haroon@iiserb.ac.in}


\begin{abstract}

Traffic sign recognition (TSR) systems deployed on embedded vehicular hardware face deployment effects that image-based benchmark evaluation does not capture, including frame-to-frame prediction instability in streaming video and load-dependent throughput under sustained operation. We present \textbf{Edge-TSR}, a continuous edge inference system for TSR on the NVIDIA Jetson Orin Nano that combines sparse detection, multi-object tracking, region-level fine-grained classification, and a lightweight track-aware temporal stabilization mechanism adding only 0.07~ms per sampled frame. Evaluating on the source videos of the INTSD benchmark, we find that the state-of-the-art TSR models lose 20--30 percentage points of mAP@50 when moving from image-based to streaming evaluation. Edge-TSR achieves the best streaming classification performance among all evaluated baselines, and temporal stabilization improves accuracy by 10.58 percentage points over per-frame inference. We characterize latency, throughput, and thermal behavior across sampling intervals and four driving conditions, showing that throughput is load-proportional ($\approx$20~ms per classified region) and that every-frame inference is limited by throughput rather than thermal throttling. A 55-minute, 26~km vehicular deployment demonstrates sustained online operation at 16.18~FPS within safe thermal limits without cloud offload. We also release the annotated 50,732-frame streaming evaluation dataset and full implementation at \href{https://anonymous.4open.science/r/Edge-TSR-29B3}{\textcolor{blue}{https://anonymous.4open.science/r/Edge-TSR-29B3}}.

\end{abstract}

\received{20 February 2007}
\received[revised]{12 March 2009}
\received[accepted]{5 June 2009}
\maketitle
\pagestyle{plain}
\thispagestyle{plain}
\section{Introduction}
Traffic sign recognition (TSR) is a safety-relevant perception function in modern vehicles: it underpins driver-assistance features such as speed-limit warning and intelligent speed assistance, and is increasingly expected to run on low-cost embedded hardware rather than in the cloud~\cite{anoop2025real, uprety2026optimizing}. A deployed TSR system does not see curated images. It sees a continuous video stream from a moving vehicle, in which each sign is visible only for a short window, appears small at a distance, and must be distinguished from visually similar classes (e.g., adjacent speed limits) under motion blur, glare, rain, and low light. Yet TSR methods are overwhelmingly developed and evaluated on static image benchmarks~\cite{stallkamp2012man, zhu2016traffic, mishra2025benchmarking, tabernik2019deep}, and their behavior under continuous on-vehicle operation remains poorly characterized. This gap is not incidental. It is a structural consequence of evaluating systems on independent image samples drawn from a stationary distribution~\cite{shankar2021image, li2020towards, koh2021wilds}, when the deployment context is a temporally correlated, non-stationary video stream subject to motion-induced degradations, variable scene complexity, and sustained thermal load on embedded hardware that benchmark protocols do not replicate~\cite{mishra2025benchmarking}.

This deployment gap is harder to close than it appears. A continuous inference pipeline operating on a moving vehicle must simultaneously satisfy constraints that are in direct tension with one another. It must sustain real-time throughput while managing the thermal envelope of embedded hardware that throttles under prolonged GPU load \cite{benoit2020impact} — a property invisible in short-duration benchmarks but dominant in vehicular operation. It must suppress the frame-to-frame prediction instability inherent to streaming video inference: motion blur, rolling-shutter artifacts, transient occlusions, and illumination changes that compound to degrade classification quality far below static-image benchmark figures~\cite{shankar2021image}. And it must do both while operating under the variable per-frame computational cost of a decoupled detection-classification architecture, where throughput is load-proportional rather than fixed — a property that creates an inverse relationship between scene complexity and system efficiency with direct consequences for deployment planning. Prior work has addressed these constraints individually — proposing efficient detectors~\cite{lin2025yolo, chen2025yolo, mishra2025benchmarking}, temporal reasoning for video~\cite{zhu2017flow, han2020mining}, or edge-optimized inference~\cite{ayachi2022edge, uprety2026optimizing, anoop2025real, zeng2020distream, fang2018nestdnn} — but rarely in combination, and not under evaluation conditions that expose the full scope of real-world deployment behavior.

The consequence is a systematic overestimation of deployed TSR performance. Accuracy reported on curated image datasets does not account for the temporal noise (frame-to-frame prediction instability: motion blur, partial occlusions, illumination variation across frames, and detector instability) introduced by streaming video; throughput reported from single forward passes does not account for thermal throttling under continuous operation; and scores on balanced, well-lit test sets do not reflect degradation under rain, low light, or brief at-speed sign exposures. The result is TSR systems that perform well in the laboratory and poorly on the road, a failure mode that matters directly when a misread speed limit cannot be corrected by a human operator in the loop.

In this work, we address these challenges through the design and characterization of \textbf{Edge-TSR}, a continuous edge inference system for TSR on resource-constrained embedded hardware. TSR is a demanding case for embedded streaming inference: signs are small, fine-grained, and briefly visible; scene complexity ranges from empty rural roads to dense urban clutter; and the system must operate continuously for the full duration of a drive. These properties expose deployment effects (temporal prediction instability, thermal sustainability, and workload-dependent throughput) that are well characterized in server-side video analytics~\cite{jiang2018chameleon, bhardwaj2022ekya} but poorly understood for single-device, on-vehicle TSR. Edge-TSR integrates a lightweight detection stage, multi-object tracking, classification, and a novel track-aware temporal stabilization mechanism that performs confidence-weighted stateful inference with hysteresis-based label locking. The system is designed around three deployment-driven objectives: maintaining sustained throughput under prolonged GPU utilization, improving temporal prediction stability in streaming inference, and managing workload-dependent computational variability on embedded hardware. Unlike conventional frame-wise TSR pipelines~\cite{tabernik2019deep, chen2025yolo, lin2025yolo, mishra2025benchmarking} and prior temporal TSR methods that operate offline without on-device constraints~\cite{zhang2022detection}, Edge-TSR explicitly treats temporal consistency and deployment sustainability as first-class system objectives rather than secondary evaluation artifacts.

We evaluate Edge-TSR under four real-world deployment conditions: dense urban traffic, rain, rural/extreme low-light driving, and out-of-distribution sign appearances. Our evaluation jointly characterizes detection and classification quality, temporal prediction stability, per-component latency, throughput, thermal envelope, and memory pressure under sustained operation. Across three independent state-of-the-art baselines, we observe a consistent 20--30 percentage points (pp) in deployed performance when transitioning from static-image evaluation to streaming video inference, demonstrating that benchmark-centric evaluation systematically overestimates real-world behavior. We further show that the proposed track-aware temporal stabilization mechanism recovers 10.58~pp classification accuracy over per-frame inference baselines at negligible computational overhead, enabling stable streaming inference under continuous operation. Finally, a 55-minute vehicular deployment over a 26~km urban and peri-urban route demonstrates sustained operation at 16.18~FPS (online) within safe thermal limits on a single Jetson Orin Nano without cloud offload.

Our findings show that image-based evaluation systematically overstates the deployed performance of embedded TSR. Our measurements are specific to TSR on the Orin Nano; however, the evaluation method (streaming, long-duration, and thermally instrumented) applies directly to other on-vehicle perception tasks, as we discuss in Section~\ref{sec:discussion}. We summarize our contributions as follows:

$\bullet$ We present \textbf{Edge-TSR}, a deployment-oriented continuous edge inference system for TSR on resource-constrained embedded hardware, evaluated under four challenging real-world  conditions.

$\bullet$ We demonstrate a consistent 20--30~pp benchmark-to-deployment performance gap across three baseline systems, showing that image-based evaluation systematically overstates deployed performance under continuous streaming inference.

$\bullet$ We introduce a lightweight (negligible overhead) track-aware temporal stabilization that improves streaming inference, recovering 10.58~pp in classification accuracy over per-frame baselines.

$\bullet$ We provide a comprehensive deployment characterization of Edge-TSR, including latency, throughput, memory usage, and thermal behavior, validated through a 55-minute vehicular deployment trial. We release our 50,732-frame streaming evaluation dataset and full system implementation.

\section{Related Work} 
\textbf{The benchmark-to-deployment gap:} The disconnect between controlled evaluation and deployment reality is a known problem in embedded AI~\cite{bianco2018benchmark, canziani2016analysis}: component-level benchmarks routinely overstate achievable system throughput by ignoring thermal throttling, memory contention, and concurrent process overhead. More recent work has highlighted similar controlled-to-field performance gaps in mobile inference systems. For example, \cite{xu2022mandheling} demonstrate that speech recognition systems evaluated under static laboratory conditions exhibit substantial degradation in real-world mobile deployment. Existing TSR and edge vision studies similarly rely primarily on curated image datasets or short-duration inference evaluation, leaving the effects of temporal instability, sustained operation, and deployment-induced workload variability largely uncharacterized. In this work, we depart from benchmark-centric evaluation and instead study continuous streaming inference under real deployment conditions on embedded hardware, jointly characterizing recognition quality, throughput, latency, and thermal behavior during sustained operation.

\textbf{Edge deployment of continuous vision pipelines:} Deploying object recognition workloads on resource-constrained edge hardware has attracted substantial attention due to growing demand from autonomous vehicles, mobile sensing, and IoT applications~\cite{han2015deep, jacob2018quantization, hinton2015distilling}. Prior work has explored hardware-aware optimization strategies including FPGA acceleration~\cite{ayachi2022edge}, lightweight inference on Raspberry Pi platforms~\cite{uprety2026optimizing}, and deployment on NVIDIA Jetson devices~\cite{anoop2025real}. While these systems improve throughput and deployment feasibility, existing studies primarily optimize individual pipeline components and rarely characterize the system-level effects that emerge under sustained real-world operation. In particular, prior work largely omits evaluation of thermal sustainability, workload-dependent throughput variability, sampling behavior, and latency overhead during continuous streaming deployment. These deployment effects, rather than isolated model accuracy, dominate the real-world behavior of continuously operating edge inference systems and are the primary focus of this paper.

\textbf{Temporal consistency in streaming inference:} Temporal reasoning has been widely studied in video understanding and object detection~\cite{wang2016temporal, tran2015learning, zhu2017flow, han2020mining}, where information propagation across frames improves prediction stability and reduces redundant inference. However, many existing approaches rely on computationally intensive techniques such as optical flow estimation or recurrent feature aggregation, making them difficult to deploy on resource-constrained embedded hardware operating under real-time constraints. Multi-object tracking approaches including SORT~\cite{bewley2016simple}, DeepSORT~\cite{wojke2017simple}, ByteTrack~\cite{zhang2022bytetrack}, BoT-SORT~\cite{aharon2022bot}, and StrongSORT~\cite{du2023strongsort} improve temporal continuity by associating detections across frames, though often with trade-offs between robustness and computational overhead. Existing edge vision pipelines typically perform frame-wise classification independently, without maintaining temporally stable per-object inference state across streaming video. In contrast, we introduce a lightweight track-aware temporal stabilization mechanism that integrates tracking with confidence-weighted stateful inference and hysteresis-based label locking to improve streaming inference stability at negligible computational overhead.

\textbf{Traffic sign recognition:} TSR has emerged as a representative workload for embedded roadside perception due to its combination of small-object detection,  classification, rapid appearance variation, and strict real-time constraints. Recent work has largely focused on adapting YOLO-family detectors and lightweight classification architectures for embedded deployment~\cite{redmon2016you, chen2025yolo, lin2025yolo, mishra2025benchmarking}. Other studies explore robustness under adverse weather, nighttime driving, and low-light conditions~\cite{sermanet2011traffic, stallkamp2012man, eykholt2018robust, uikey2024indian, li2017perceptual}. However, most prior TSR evaluations are conducted on static image datasets or extracted frames rather than continuous streaming video. As a result, existing studies rarely characterize the degradation in detection and classification performance that emerges under temporally correlated deployment conditions. In Section~\ref{sec:results}, we systematically quantify this benchmark-to-deployment gap across three independent baselines.

\section{System Design}
\label{sec:method}


\subsection{Design Constraints and Trade-offs}
The design of Edge-TSR is governed by three system-level tensions that emerge during continuous deployment on embedded hardware.  \textbf{(1) Throughput vs. accuracy under sparse sampling:} per-frame inference sustains only 13.25~FPS on the Jetson Orin Nano (Section~\ref{sec:results}), below the input frame rate. Sparse sampling restores real-time operation but leaves temporal gaps, which the stabilization module bridges so that recognition quality does not degrade in proportion to inference frequency. \textbf{(2) Temporal stability vs. responsiveness to genuine sign changes:} the hysteresis locking mechanism is deliberately asymmetric, requiring strong sustained contradictory evidence to change a locked real-class label while applying a more permissive escape condition for the \textit{unknown} class where suppression is less critical. \textbf{(3) Classification quality vs. edge memory budget:} applying the high-capacity  classifier only to localized regions of interest — rather than full frames — bounds its cost by the number of active detections per frame, making it feasible within the Jetson's unified 8~GB memory pool.
 These deployment constraints jointly motivate the system architecture described below. Section~\ref{sec:results} evaluates the contribution of each design choice through controlled ablation and deployment experiments.
 
\subsection{System Overview}
\begin{figure*}[]
    \centering
    \includegraphics[width=0.75\linewidth]{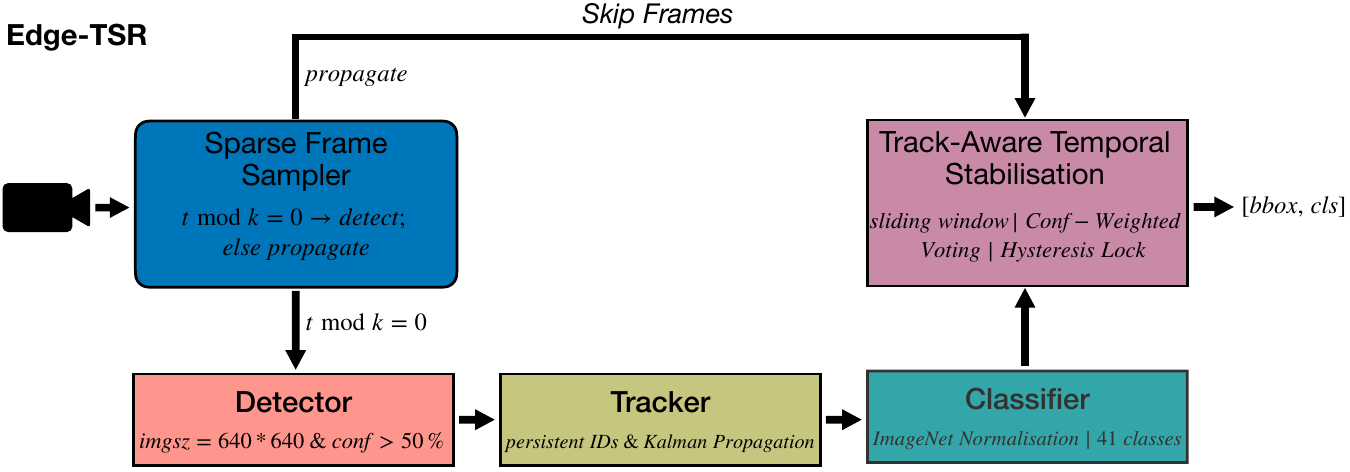}
    \caption[Overview of the Edge-TSR pipeline]{Overview of the Edge-TSR continuous edge inference system. Detection is performed every $k$ frames, with tracking-based state propagation on intermediate frames. A track-aware temporal stabilization layer combines classification outputs through confidence-weighted voting and hysteresis to produce temporally stable $[\textit{bbox}, \textit{cls}]$ outputs.
    }
    
    \label{fig:pipeline}
\end{figure*}

Edge-TSR processes a continuous video stream through four stages: sparse detection, tracking-based state propagation, fine-grained classification, and track-aware temporal stabilization (Figure~\ref{fig:pipeline}).

For a video stream $V=\{f_1,f_2,\ldots,f_n\}$, where $f_i$ denotes $i^{th}$ video frame, full detection inference is performed only on frames satisfying $t \bmod k = 0$, where $k$ is the sampling interval. On non-sampled frames, tracked object states are propagated from the previous detection frame, and previously inferred labels are carried forward. The pipeline produces per-frame outputs with bounding boxes, persistent track identities, and stabilized class labels.
 
\subsection{Detection and Tracking} 

Traffic sign localization is performed using a lightweight real-time detector optimized for embedded deployment. The detector operates at fixed input resolution, providing sufficient spatial detail for sign localization while remaining within the memory and latency constraints of the Jetson Orin Nano. Detection is class-agnostic: the detector localizes candidate sign regions, while fine-grained semantic recognition is delegated to a downstream region-level classifier. Our implementation uses YOLOv8 \cite{yolov8_ultralytics} due to its favorable accuracy-to-latency trade-off, native TensorRT support, and efficient integration with the tracking subsystem on Jetson.

\textbf{Multi-object tracking:} To maintain temporal continuity across frames, Edge-TSR integrates a lightweight tracking layer that associates detections over time and assigns persistent identities to active objects. For each frame $f$, the tracker maintains a set of active tracklets $\mathcal{T} = \{(B_i^f, \text{ID}_i)\}$, where $B_i^f = (x_1, y_1, x_2, y_2)$ denotes the bounding box of object $i$ and $\text{ID}_i$ is its persistent identity. We use ByteTrack~\cite{zhang2022bytetrack}, which exploits both high- and low-confidence detections to preserve track continuity under partial occlusion and transient detector instability. On intermediate frames where detection is skipped, the tracker propagates object states using Kalman filter prediction without invoking the detector. Integrating tracking directly into the inference pipeline serves two   objectives: (i) enabling temporal association of classification outputs under a shared object identity, which supports the stabilization layer described in Section~\ref{sec:temporal-stab}, and (ii) reducing redundant detector invocations across temporally correlated frames during sparse scheduling.

\textbf{Minimum size filtering:} Detections with bounding box dimensions below $d_{min} \times d_{min}$ pixels are discarded prior to classification. This removes highly distant or truncated detections for which crop quality is insufficient for reliable fine-grained recognition and reduces downstream computation on visually ambiguous regions.

\subsection{Fine-Grained Classification}

Each detected Region of Interest (RoI) is processed using a high-capacity fine-grained classifier trained to discriminate among 41 traffic sign categories on INTSD \cite{mishra2025benchmarking}. The classifier operates only on localized tracked regions rather than full frames, allowing computational cost to scale with active scene complexity rather than image resolution. We instantiate this component using ResNet-50 \cite{he2016deep}, selected based on empirical evaluation on the Jetson Orin Nano and its favorable balance between representational capacity and deployment feasibility. A detailed comparison with alternative architectures is provided in Appendix~\ref{sec:appendix}. The classifier produces a probability distribution over classes via a softmax output layer, and the predicted label is:
\[
\hat{y} = \arg\max_{c} \; P(c \mid x)
\]
where $c$ denotes the class and $x$ denotes the detected signboard.

\textbf{Contextual padding:}  Detector bounding boxes are expanded by a padding factor $p$ in both spatial dimensions before cropping:
\begin{equation*}
    dx = \lfloor (x_2 - x_1) \cdot p \rfloor, \quad
    x_1' = \max(0,\, x_1 - dx), \quad
    x_2' = \min(W,\, x_2 + dx)
\end{equation*}

and analogously for the vertical dimension. This expansion serves two purposes: (i) it provides the classifier with contextual border information (sign background, mounting structure, and partially truncated sign regions) that aids discrimination, and (ii) it reduces sensitivity to localization jitter from the detector. Crops are resized to a fixed dimension and normalized using ImageNet channel statistics ($\mu = [0.485, 0.456, 0.406]$, $\sigma = [0.229, 0.224, 0.225]$).

\textbf{Geometric pruning for visual noise:} Roadside environments contain visually confusable objects (commercial advertisements, branded stickers, and informational placards) that may trigger detector responses and be misclassified. To suppress such false positives, we introduce a geometric pruning rule: any detection classified as advertisement with bounding box width or height below $d_{adv}$ pixels is reassigned to the unknown class. This rule encodes the prior that genuine traffic signs occupy a minimum physical footprint when projected onto the image plane at operationally relevant distances, while advertisement-class responses on small detections are more likely to correspond to roadside visual noise.

The decoupled design ensures that the classifier's high capacity and large input resolution do not impose latency at the detection stage. Classification is invoked only for active detections, and its cost is bounded by the number of tracked signs in the scene rather than the full frame resolution.

\subsection{Track-Aware Confidence-Weighted Temporal Stabilization}
\label{sec:temporal-stab}

Streaming inference under real driving conditions introduces significant temporal noise in per-frame classification outputs. Sources of noise include: motion blur from vehicle movement, rolling-shutter artifacts, partial occlusions as the vehicle passes a sign, rapid illumination changes (oncoming headlights), and detector instability manifesting as transient missed detections or slightly shifted bounding boxes. These effects are exacerbated under sparse inference, where inference is not performed on every frame.

To address label flickering while avoiding the overhead of per-frame temporal models, we propose a lightweight track-aware confidence-weighted temporal stabilization module. The module operates independently per tracked object and introduces negligible computational cost relative to detection and classification.

\textbf{Memory buffer:} For each tracked object $i$ with identity $\text{ID}_i$, we maintain a sliding window memory buffer of fixed length $T$ ($T=5$):
\[
M_i = \{(y_j, s_j)\}_{j=t-T+1}^{t}
\]
where $y_j$ is the classifier's predicted label and $s_j \in [0,1]$ is the associated softmax confidence at frame $j$. The buffer stores only frames on which classification was invoked, i.e., detection frames under the sparse sampling schedule.

\textbf{Majority voting with confidence-gated locking:} The candidate label is the most frequent label within the window,
\[
\tilde{y}_i = \arg\max_{c} \sum_{(y_j, s_j) \in M_i} \mathds{1}[y_j = c],
\]
while confidence enters through the locking condition below, which additionally requires the mean confidence of the majority label to exceed $\delta$. A label that appears frequently but with low confidence can therefore be output as a candidate, but is never locked; likewise, \textit{unknown} is never locked from the candidate state.

\textbf{Hysteresis-based label locking:} Confidence-weighted voting is sufficient to suppress high-frequency noise but does not prevent slower-timescale label oscillations caused, for example, by a partially occluded sign alternating between two visually similar classes over several seconds. We introduce a hysteresis mechanism that promotes a candidate label to a locked state when evidence reaches a stability threshold, and requires strong contradictory evidence to permit a subsequent state transition. Figure~\ref{fig:hysteresis} provides a detailed illustration of the proposed mechanism. The stabilization module operates as a two-state machine per track:

\begin{figure}[t]
    \centering
    \includegraphics[width=0.95\columnwidth]{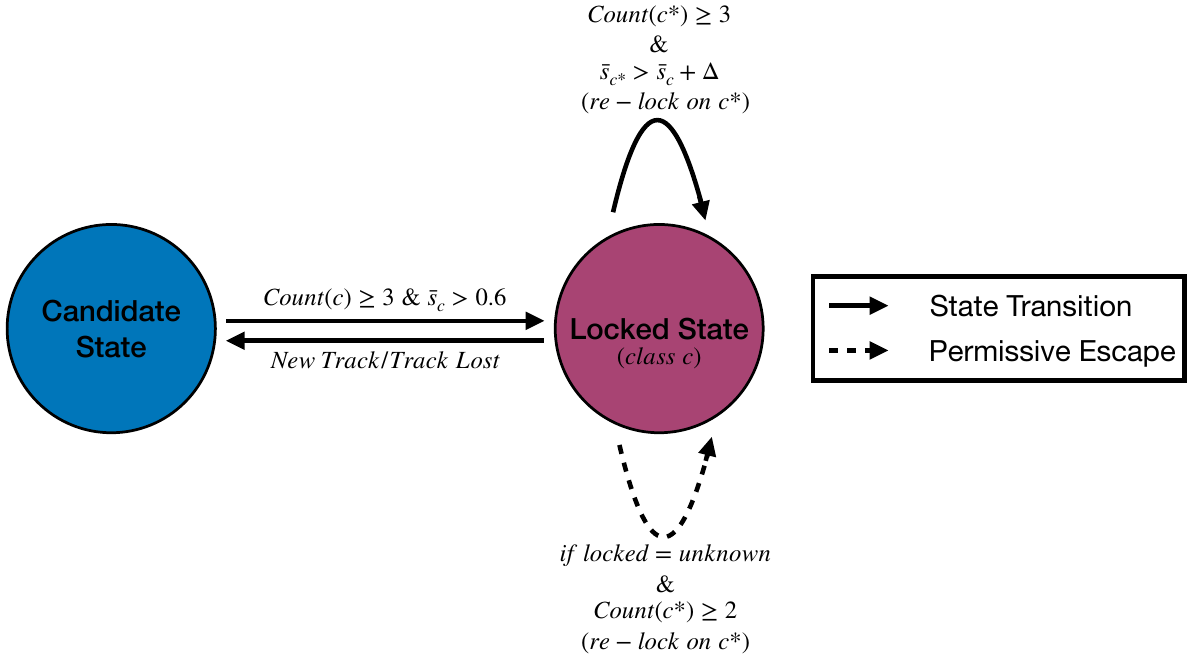}
    \caption[State machine of the track-aware temporal stabilization module]{State machine of the track-aware temporal stabilization layer. Tracks transition from a \textit{Candidate} state to a \textit{Locked} state based on confidence and temporal consistency. Locked labels are retained across frames and updated only when sufficient contradictory evidence is observed.
    }
    \label{fig:hysteresis}
\end{figure}

\textit{\textbf{ (i) Candidate state.}} When a track has no locked label, per-frame predictions from confidence-weighted voting are used directly. A label becomes eligible for locking when:
\[
\text{Count}(c) \geq \tau \quad \text{and} \quad \overline{s}_{c} > \delta
\]
where $\text{Count}(c)$ is the frequency of the majority label within $M_i$ and $\overline{s}_{c}$ is the mean softmax confidence over all observations of $c$ in the window. The threshold values are determined empirically.

\textit{\textbf{(ii) Locked state.}} Once a real-class label is locked, it is output for all subsequent frames regardless of single-frame classifier outputs. Label oscillations arising from transient noise are suppressed entirely. \textit{Unknown}-class detections within the window are ignored: a locked real label is not displaced by uncertain predictions.

\textit{\textbf{Escape condition:}} To prevent misclassification persistence — arising, for instance, when an out-of-distribution object is initially locked to an incorrect label — the module permits a state transition from locked label $c$ to a new candidate label $c^{*}$ (re-lock on $c^{*}$) only when the new evidence satisfies a confidence premium:

\[
\text{Count}(c^{*}) \geq \tau \quad \text{and} \quad \overline{s}_{c^{*}} > \overline{s}_{c} + \Delta
\]

where $\Delta$ is the confidence margin. This asymmetric escape condition ensures that state changes occur only when there is sustained, high-confidence contradictory evidence, and not in response to momentary artifacts. A special, more permissive escape rule applies when the locked label is \textit{unknown}: any majority non-unknown label appearing with $\text{Count} \geq \tau -1$ immediately supersedes the unknown lock, as suppression of unknown is less safety-critical than suppression of real class changes. The module's computational cost is $O(T)$ per track per frame, where $T=5$, making it negligible relative to the inference pipeline.

\subsection{Periodic Sparse Frame Sampling}

Performing full detection and classification on every frame is computationally expensive and thermally unsustainable under continuous operation on embedded hardware. Beyond raw computational cost, exhaustive per-frame inference leads to thermal accumulation and subsequent CPU/GPU frequency scaling, producing non-linear throughput degradation during long-duration deployment. Edge-TSR therefore adopts a periodic sparse inference strategy in which full detection inference is invoked every $k^{th}$ frame, corresponding to a duty cycle of (100/$k$)\%. On intermediate frames, the system reuses tracked object states propagated by the tracking layer and retains the most recent stabilized classification labels.

This design is motivated by the observation that consecutive driving frames exhibit substantial temporal redundancy: object identity changes slowly relative to frame rate, and object displacement over short intervals remains small relative to the contextual padding applied during region-level classification. Sparse scheduling therefore reduces computational load and thermal pressure while preserving stable inference behavior through temporal state propagation and stabilization.

The choice of $k=3$ is determined empirically and evaluated in Section~\ref{sec:results}. Combined with the temporal stabilization layer, periodic sparse inference enables sustained real-time throughput under continuous deployment while avoiding the thermal instability associated with exhaustive per-frame inference.


\section{Dataset and Evaluation Environment} \label{sec:data}
\subsection{Training Data: INTSD Images}

We train the on the Indian Nighttime Traffic Sign Dataset (INTSD)~\cite{mishra2025benchmarking}, the only public fine-grained TSR dataset collected under nighttime and adverse-weather conditions on Indian roads. Its visual noise (motion blur, lens glare, partial occlusions, and truncated signs at frame boundaries) matches the crop quality of a detector on a moving vehicle, reducing the train-deployment domain gap. INTSD contains 41 classes with 14,044 annotated instances. We use Fold~0 of the standard five-fold split, which generalizes best to streaming video, and apply YOLO's default augmentations (mosaic, mixup, multi-scale training) plus translation and rotation to capture spatial and viewpoint variation in real driving scenario.

\subsection{Evaluation Data: INTSD Source Videos}

Existing TSR benchmarks, including INTSD~\cite{mishra2025benchmarking}, are static images curated from driving recordings and do not reflect the temporal dynamics, scene density, and environmental diversity of continuous operation. To separate streaming effects from domain shift, our in-domain evaluation videos are the raw source recordings from which the INTSD images were curated~\cite{mishra2025benchmarking}. Image-based and streaming evaluation therefore share the same camera, routes, scenes, and conditions, differing only in protocol: INTSD discarded frames with severe glare or motion blur, whereas streaming evaluation processes every frame. The out-of-distribution video was recorded separately and does not overlap with INTSD or its class taxonomy.


\textbf{Dataset composition.} The evaluation set comprises four videos, each corresponding to a distinct operating condition:
\textbf{(i) Dense urban traffic.} Recorded under normal nighttime conditions in high-traffic urban zones with high sign density, frequent occlusion from vehicles, and background clutter from roadside commercial signage. This condition establishes the operational performance baseline.
\textbf{(ii) Rain.} Recorded during active precipitation. This condition introduces lens occlusion from water droplets, specular highlights from wet road surfaces, contrast reduction, and intermittent detector confidence suppression. 
\textbf{(iii) Rural and dark (no street lighting).} Recorded on rural roads in the absence of street lighting, with illumination provided exclusively by the vehicle's headlights. This condition represents the extreme of the nighttime operating envelope, with high dynamic range, sign reflectivity variation, and low background contrast.
\textbf{(iv) Out-of-distribution (OOD).} Recorded in locations containing both (a) in-vocabulary sign categories rendered in non-standard physical forms — differing shapes, materials, or typographic styles not present in INTSD, and (b) sign categories entirely absent from the 41-class training vocabulary. This condition stress-tests the system's behavior at the boundary of its closed-world classification assumption.

\textbf{Annotation.} All 50,732 frames are manually annotated in CVAT \cite{cvat} following the protocols of \cite{mishra2025benchmarking} and \cite{ertler2020mapillary}, with bounding boxes and class labels for every visible sign, including partially occluded and motion-blurred instances. Of these, 24,721 frames contain signs, with 62,707 annotated instances in total. Each video was reviewed frame by frame three times to minimize labeling errors. Signs below a minimum visibility are excluded so that the detector is not penalized for physically unresolvable instances. To our knowledge, this is the first public streaming evaluation dataset for continuous edge inference under adverse roadside conditions on Indian roads.

\begin{table}[t]
\centering
\caption{Dataset statistics across different evaluation scenarios. The average bounding box area is given in $pixels^{2}$.}
\label{tab:scenario_stats} 
\resizebox{\columnwidth}{!}{
\begin{tabular}{lcccccc}
\toprule
\textbf{Scenario} &
\begin{tabular}[c]{@{}c@{}}\textbf{Total}\\\textbf{Frames}\end{tabular} &
\begin{tabular}[c]{@{}c@{}}\textbf{Labeled}\\\textbf{Frames}\end{tabular} &
\begin{tabular}[c]{@{}c@{}}\textbf{Sign}\\\textbf{Instances}\end{tabular} &
\begin{tabular}[c]{@{}c@{}}\textbf{Avg. Inst.}\\\textbf{/Frame}\end{tabular} &
\begin{tabular}[c]{@{}c@{}}\textbf{Max.}\\\textbf{Inst.}\end{tabular} &
\begin{tabular}[c]{@{}c@{}}\textbf{Avg. Box}\\\textbf{Area}\end{tabular} \\
\midrule
Dense & 27030 & 19290 & 53231 & 2.76 & 11 & 13231.92 \\
Rural/Dark & 7042 & 914 & 1127 & 1.23 & 5 & 4684.00 \\
OOD & 4413 & 1895 & 2181 & 1.15 & 4 & 5499.35 \\
Rain & 12247 & 2622 & 6168 & 2.35 & 7 & 13199.34 \\
\bottomrule
\end{tabular}
}
\end{table}

\textbf{Dataset statistics.} Table~\ref{tab:scenario_stats} summarizes each scenario. Dense traffic contains the highest object density (2.76 instances/frame, up to 11 per frame) and the largest average bounding box area (13,231~px$^2$), while rural and OOD settings are considerably sparser and contain smaller, more distant signs (4,684 and 5,499~px$^2$ respectively). These differences directly influence detection difficulty and explain the performance variation reported in Section~\ref{sec:results}. 

\section{System Evaluation} \label{sec:system_evaluation}

\subsection{Evaluation Scenarios}

The four scenarios (Section~\ref{sec:data}) each stress a different system property.
\textbf{(i) Dense urban traffic} imposes the highest load (up to 11 signs per frame) and thus characterizes worst-case throughput.
\textbf{(ii) Rain} tests whether temporal stabilization maintains correct labels when precipitation suppresses detector confidence.
\textbf{(iii) Rural and dark} isolates the nighttime detection bottleneck from the classification bottleneck (Section~\ref{sec:results}).
\textbf{(iv) OOD} probes two generalization boundaries, (a) intra-class distribution shift, where known sign categories appear in non-standard physical forms, and (b) true open-set failure, where sign categories have no in-distribution equivalent in the 41-class training vocabulary. These are evaluated jointly but interpreted separately in Section \ref{sec:results}.

All scenarios use one fixed configuration ($k=3$, confidence threshold 0.5, padding $p=0.20$) on the Jetson Orin Nano, with no scenario-specific tuning.

\subsection{Hardware Platform}



All controlled evaluations and the live trial run on the \textbf{NVIDIA Jetson Orin Nano} (8~GB), which provides up to 67 sparse (33 dense) INT8 TOPS within a 7--25~W power envelope at a developer-kit price of USD~399~\cite{nvidia_orinnano}. It combines a 6-core Arm Cortex-A78AE CPU with an NVIDIA Ampere GPU (1024 CUDA cores, 32 Tensor Cores). CPU and GPU share a single 8~GB LPDDR5 pool, which constrains concurrent allocation by the inference engine, the OpenCV capture backend, and the OS. We use the MAXN SUPER power mode, which permits the highest sustained clocks. Full specifications are mentioned in the Appendix~\ref{sec:appendix}.

\textbf{Choice of platform.} We target the compute, power, and cost class of a standalone front-camera TSR module. In production vehicles, TSR is a base-ADAS function typically served by a single windshield camera and a low-cost SoC such as the Mobileye EyeQ6 Lite (5 INT8 TOPS)~\cite{mobileye_portfolio, mobileye_eyeq}; centralized higher-capacity platforms run multi-task autonomy stacks in which TSR receives only a fraction of the compute. The Orin Nano (67 INT8 TOPS) thus exceeds the compute nominally available to production TSR while staying within a mass-market cost and power envelope.

\begin{figure}[htbp]
    \centering
    \includegraphics[width=0.9\columnwidth]{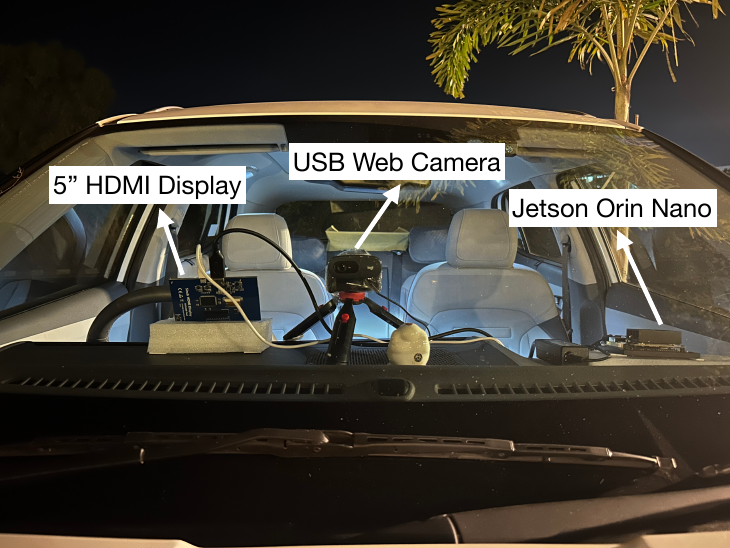}
    \caption{Real-world deployment setup. The NVIDIA Jetson Orin Nano (right) is mounted on the vehicle dashboard alongside a 5-inch HDMI display (left) for live pipeline visualization. A Logitech USB webcam is mounted on a tripod at the center of the dashboard for forward-facing video capture.}
    \label{fig:car}  
\end{figure}

\subsection{Real-World Deployment Setup}


The Jetson is mounted on the dashboard with a 5-inch HDMI display showing the live annotated feed (boxes, labels, confidences, and FPS) for verification (Figure~\ref{fig:car}). Following \cite{anoop2025real}, a Logitech C270 HD webcam on a dashboard tripod at the center of the windscreen captures forward-facing 720p video at 30~FPS. A single runtime on the Jetson handles capture, sparse inference scheduling, tracking, stabilization, visualization, and metric logging without cloud offload, reflecting the minimal footprint of a production edge system.

\subsection{Implementation Details}

\textbf{Training protocol.} The YOLOv8 detector is fine-tuned on INTSD Fold~0 with the augmentations described in Section~\ref{sec:data}, and the ResNet-50 classifier is trained with cross-entropy loss from ImageNet-pretrained initialization.

\textbf{Inference pipeline.} Edge-TSR is implemented in Python with PyTorch 2.10 and OpenCV 4.13. YOLOv8 is loaded via the Ultralytics API~\cite{yolov8_ultralytics} and run on the Jetson GPU in FP32 without TensorRT, reflecting off-the-shelf deployment rather than hardware-specific tuning. ByteTrack runs through the Ultralytics tracking interface with default hyperparameters (\textit{bytetrack.yaml}), and ResNet-50 runs in FP32. Zero-copy buffers between OpenCV capture and PyTorch minimize host-device transfer overhead.

System metrics include FPS, per-component latency, CPU/GPU utilization, temperature, and memory usage. They are sampled at every frame using \textit{psutil} and the Jetson hardware monitoring interface, and logged alongside prediction outputs for post-hoc analysis. All controlled experiments are conducted over five independent runs and we report the mean of the resulting metrics.

\subsection{Evaluation Metrics}

We categorize evaluation metrics into four groups: (i) system-level metrics, which characterize computational and thermal sustainability on edge hardware; (ii) temporal consistency, which quantifies resistance to label flickering under streaming inference; (iii) detection metrics, which assess localization quality independently of class label; and (iv) classification metrics, which assess fine-grained label correctness on RoIs. 

\subsubsection{System-Level Metrics}
End-to-end \textbf{throughput (FPS)} is measured from raw image loading to final predictions, including all preprocessing, forward pass, and post-processing~\cite{mishra2025benchmarking}. In controlled evaluations, frames are decoded from recorded video and processed as fast as possible, so FPS denotes \emph{offline} throughput, averaged over five runs. In the real-world deployment, frames are captured from a live camera, so FPS denotes \emph{online} throughput. We adopt FPS $\geq 15$ as the threshold for real-time operational viability~\cite{wagner2009real, redmon2016you}. \textbf{Latency.} Under sampling interval $k$, a \emph{sampled} frame ($t \bmod k = 0$) executes detection with tracker association (Det+Trk, a single call in our implementation), classification of each detected region (Cls), and temporal stabilization (Stab); a \emph{non-sampled} frame only propagates the most recent boxes and stabilized labels without inference. Det+Trk, Cls (per region), and Stab are reported as mean per-invocation wall-clock time. We denote by $L_s$ and $L_p$ the mean end-to-end latency of sampled and non-sampled frames, respectively, and report the amortized per-frame latency: $L_{\text{avg}} = \bigl(L_s + (k-1)L_p\bigr)/k = 1000/\text{FPS}$~ms. \textbf{CPU/GPU utilization} and \textbf{thermal envelope} (average and maximum CPU, GPU, and junction (Tj) temperature in $\degree$C) are logged continuously with \texttt{jtop}; on the Jetson Orin Nano, thermal throttling reduces clock frequencies and degrades throughput, so the thermal envelope characterizes pipeline sustainability under continuous operation. \textbf{Memory utilization} is reported as average and peak RAM and GPU memory occupancy as a percentage of the unified 8~GB pool.


\subsubsection{Temporal Consistency}
To measure the system's resistance to label flickering — a failure mode specific to streaming video inference that does not appear in image-based evaluations — we define a \textbf{consistency} metric. For each track $i$ with label sequence $\{y_t\}_{t=1}^{L_i}$, we compute the fraction of consecutive frame-pairs in which the label does not change:

\begin{equation}
    \mathrm{Consistency}_i = 
    \frac{1}{\sum_{k} (T_k - 1)} \sum_{k} \sum_{t=2}^{T_k} \mathbb{1}\left[y_t^{(k)} = y_{t-1}^{(k)}\right]
\end{equation}

Overall consistency is averaged across all tracks with length $L_i \geq 5$, a minimum length threshold that excludes transient detections from contributing to the metric:
\begin{equation}
    \mathrm{Consistency} = 
    \frac{1}{|\mathcal{T}|} \sum_{i \in \mathcal{T}} \mathrm{Consistency}_i
\end{equation}
where $\mathcal{T}$ is the set of qualifying tracks. The complement, $\mathrm{Flip\ Rate} = 1 - \mathrm{Consistency}$, measures the proportion of consecutive observations where the system changes its label for the same physical sign. High consistency does not imply correctness — a system that consistently outputs the wrong label scores 100\% consistency — and should therefore be interpreted jointly with the classification
metrics defined in Section~\ref{sec:metrics:classification}. Its utility is in quantifying label oscillation as an independent failure mode: a system with high accuracy but low consistency produces unreliable outputs that would cause instability in downstream navigation or driver alert modules.

\subsubsection{Detection Metrics}

We report \textbf{mAP@50} and \textbf{mAP@50:95} following the standard PASCAL VOC and COCO evaluation protocols~\cite{everingham2010pascal}, measuring localization quality under lenient and strict IoU thresholds respectively.

\subsubsection{Classification Metrics}
\label{sec:metrics:classification}
Classification is evaluated on stabilized label outputs from the temporal module rather than raw per-frame classifier outputs, ensuring metrics reflect deployed system behavior. For each class $c$, a true positive ($\mathrm{TP}_c$) requires both correct localization (IoU $\geq 0.5$) and correct label assignment. We report \textbf{macro-averaged precision (A.P.)} and \textbf{macro-averaged recall (A.R.)}, which assign equal weight to each class regardless of frequency — critical in our setting because rare sign categories (e.g., \textit{cattle crossing}, \textit{school zone}) are safety-critical despite low instance counts. \textbf{Classification accuracy (Acc.)} measures the fraction of all ground-truth instances assigned the correct label:

\begin{equation}
    \mathrm{Accuracy} = \frac{\displaystyle\sum_{c=1}^{C}
    \mathrm{TP}_c}{N_{\mathrm{gt}}}
\end{equation}

where $N_{\mathrm{gt}}$ is the total number of ground-truth instances. Unlike A.P.\ and A.R., accuracy is instance-weighted and reflects aggregate system correctness including the effect of class imbalance.

\subsection{Baselines}
Baselines span the main architectural families used for TSR and edge detection: a two-stage region-proposal detector (Faster R-CNN~\cite{ren2015faster}), a lightweight single-stage mobile detector (SSDLite~\cite{sandler2018mobilenetv2}), TSR-specialized single-stage detectors (YOLO-TS~\cite{chen2025yolo} and the low-light variant YOLO-LLTS~\cite{lin2025yolo}), an edge-deployed TSR system on Jetson hardware (Anoop et al.~\cite{anoop2025real}), and a CLIP-based multimodal recognition model (LENS-Net~\cite{mishra2025benchmarking}). Together, these cover two-stage and single-stage detection, mobile-optimized and TSR-specific designs, and vision-only and vision-language recognition. All baselines are run frame-wise on the same streaming videos and the same Jetson Orin Nano, so differences reflect architecture rather than evaluation conditions.


\begin{table*}[t]
    \centering
    \footnotesize
    \caption[Comparison of models on INTSD and video-based evaluation settings]{Comparison of baseline systems under image-based and streaming video evaluation. All systems experience substantial performance degradation when transitioning from benchmark-style image evaluation to deployment-oriented video evaluation. The reported FPS is offline end-to-end throughput. $^\dag$ LENS-Net could not be executed at a practical deployment rate on the Jetson Orin Nano GPU due to its computational and memory requirements. Consequently, all reported LENS-Net results were obtained using CPU-based inference. $^\S$ YOLO-TS augmented with Edge-TSR pipeline components.}
    \label{table:results}
    \setlength{\tabcolsep}{6pt}
    \resizebox{0.80\textwidth}{!}{
    \begin{tabular}{p{3cm}cccccc}
    \toprule
    \textbf{Model} & \multicolumn{2}{c}{\textbf{Detection (\%)}} & \multicolumn{3}{c}{\textbf{Classification (\%)}} & \textbf{FPS} \\
    \cmidrule(lr){2-3} \cmidrule(lr){4-6} 
    & \textbf{mAP@50} & \textbf{mAP@50:95} & \textbf{A.P.} & \textbf{A.R.} & \textbf{Acc.} & \textbf{(end-to-end)} \\
    \midrule
    
    \multicolumn{7}{c}{\textit{Image-based Evaluation}} \\
    \midrule
    YOLO-LLTS \cite{lin2025yolo} & 90.00 & 73.65 & 68.28 & 58.13 & 83.89 & 56.30 \\
    YOLO-TS \cite{chen2025yolo} & 90.82 & 77.27 & 75.08 & 74.43 & 89.70 & 63.36 \\
    LENS-Net \cite{mishra2025benchmarking} & 94.46 & 81.56 & 85.09 & 79.32 & 88.38 & 33.87 \\
    \midrule
    \multicolumn{7}{c}{\textit{Video-based Evaluation}} \\
    \midrule
    SSDLite \cite{sandler2018mobilenetv2} & 9.32 & 6.29 & 24.79 & 1.54 & 9.31 & 39.79 \\
    Faster R-CNN \cite{ren2015faster} & 49.57 & 27.18 & 42.10 & 42.32 & 52.42 & \textbf{43.65} \\
    YOLO-LLTS \cite{lin2025yolo} & 60.50 & 41.10 & 58.55 & 46.49 & 63.03 & 16.13 \\
    Anoop et al. \cite{anoop2025real} & 60.96 & 43.47 & 42.24 & 48.29 & 65.21 & 7.72 \\
    LENS-Net \cite{mishra2025benchmarking} & 65.73 & 47.40 & 64.61 & 58.84 & 68.59 & $0.44^\dag$ \\
    YOLO-TS \cite{chen2025yolo} & 70.48 & \textbf{48.32} & 58.58 & 66.92 & 72.90 & 38.90 \\
    YOLO-TS + Edge-TSR$^\S$ & 55.84 & 35.47 & 67.49 & 57.99 & 66.98 & 38.13 \\
    \midrule
    \textbf{Edge-TSR} & \textbf{73.12} & 45.79 & \textbf{75.23} & \textbf{72.10} & \textbf{75.33} & 31.63 \\
    \bottomrule
    \end{tabular}
    } 
\end{table*}

\section{Results}
\label{sec:results} 
The evaluation is around four questions that collectively characterize the system's deployability: (\textbf{RQ1}) How does the full pipeline compare to prior TSR approaches under realistic streaming video conditions? (\textbf{RQ2}) How does detection and recognition performance vary across the environmental conditions encountered in real driving? (\textbf{RQ3}) What are the system-level trade-offs of sparse frame sampling under continuous edge operation? (\textbf{RQ4}) How does the system behave under unconstrained real-world vehicular deployment? Where applicable, we distinguish between image-based evaluation — which reflects the conditions of prior benchmark work, and video-based evaluation — which reflects deployed system behavior — to make the image-to-video performance gap explicit.

\subsection{Comparison with Prior Work (RQ1)}

Table~\ref{table:results} compares Edge-TSR against three recent state-of-the-art TSR models: YOLO-LLTS \cite{lin2025yolo}, YOLO-TS \cite{chen2025yolo}, and LENS-Net \cite{mishra2025benchmarking}, evaluated under both image-based (INTSD) and video-based protocols on our video dataset. We report detection quality (mAP@50, mAP@50:95) and classification quality (A.P., A.R., and Acc.) separately, as these measure complementary properties of the pipeline.

\textbf{Finding 1: Image-based benchmarks overstate deployed performance by 20--30~pp.} The most consequential result in Table~\ref{table:results} is not the relative ranking of models but the magnitude and consistency of the performance drop from image-based to video-based evaluation across all three baselines. YOLO-LLTS drops from 90.00\% to 60.50\% mAP@50 (29.50~pp relative degradation); YOLO-TS drops from 90.82\% to 70.48\%; LENS-Net, the strongest image-based system at 94.46\% mAP@50, drops to 65.73\%. Classification follows the same pattern: A.P. drops by 9.73\%, 16.50\%, and 20.48\% for YOLO-LLTS, YOLO-TS, and LENS-Net respectively. This gap is consistent across three independent baselines, suggesting that deployment-induced degradation is not specific to a single architecture.

\textbf{Finding 2: Edge-TSR achieves the best streaming classification at competitive detection quality.} Under video-based evaluation, Edge-TSR achieves the highest mAP@50 (73.12\% vs.\ 70.48\% for YOLO-TS), while its mAP@50:95 trails YOLO-TS by 2.53~pp (45.79\% vs.\ 48.32\%). On classification, Edge-TSR outperforms all baselines: A.P. 75.23\% versus 64.61\% for LENS-Net (+10.62~pp), A.R. 72.10\% versus 66.92\% for YOLO-TS (+5.18~pp), and accuracy 75.33\% versus 72.90\% for YOLO-TS (+2.43~pp). LENS-Net, the strongest classifier under image-based evaluation (A.P. 85.09\%), falls behind Edge-TSR by over 10~pp under streaming evaluation, suggesting that image-benchmark rankings do not carry over to continuous video. The lower mAP@50:95 reflects a trade-off of the decoupled detection-classification architecture: the class-agnostic detector does not benefit from fine-grained supervision, a localization penalty we accept in exchange for the classification gains above. Finding~3 examines which components account for these gains.

\textbf{Finding 3: Temporal stabilization alone does not explain Edge-TSR's advantage.} To isolate the stabilization layer’s contribution, we add Edge-TSR's tracking, temporal stabilization, and filtering rules to the strongest baseline, keeping YOLO-TS's backbone and sampling interval ($k=3$; YOLO-TS + Edge-TSR$^\S$ in Table~\ref{table:results}). Stabilization raises YOLO-TS's classification precision (A.P. 58.58 $\rightarrow$ 67.49\%) but lowers recall and accuracy (A.R. 66.92 $\rightarrow$ 57.99\%, Acc. 72.90 $\rightarrow$ 66.98\%). Detection drops (mAP@50 70.48 $\rightarrow$ 55.84\%) for two reasons: Edge-TSR's small-advertisement rule relabels 13.4\% of YOLO-TS's boxes as \textit{unknown}, which class-aware mAP counts as errors, and the tracker does not output 7.4\% of YOLO-TS's detections, predominantly low-confidence ones belonging to tracks not yet confirmed. Edge-TSR still outperforms this variant on all classification metrics (A.P. 75.23\%, A.R. 72.10\%, Acc. 75.33\%), whereas without stabilization its accuracy (64.75\%, \textit{No Memory} in Section~\ref{sec:ablation}) falls below that of per-frame YOLO-TS (72.90\%). The gains therefore arise from combining the decoupled detector-classifier design with stabilization, rather than from stabilization alone.

\subsection{Performance Across Driving Scenarios (RQ2)} 

\begin{table*}[t]
\centering
\caption[Performance of Edge-TSR across diverse real-world driving scenarios]{Performance across diverse real-world driving scenarios. All metrics are computed on the deployed edge system under streaming video conditions. Please refer to Appendix \ref{sec:appen_qualitative_res} for qualitative results. Abbreviations Det., Trk., and Cls. correspond to detector, tracker, and classification latencies (ms), respectively.}
\label{tab:scenario_performance}
\resizebox{0.8\textwidth}{!}{
\begin{tabular}{lccccccccc}
\toprule
\textbf{Scenario} & \textbf{mAP@50} & \textbf{mAP@50:95} & \textbf{A.P.} & \textbf{A.R.} & \textbf{Acc.} & \textbf{FPS} & \textbf{Det. + Trk.} & \textbf{Cls.} \\
\midrule
Dense & 73.12 & 45.79 & 75.23 & 72.10 & 75.33 & 31.63 & 36.77 & 21.85  \\
Rain & 61.20 & 37.81 & 54.14 & 72.70 & 68.64 & 37.40 & 32.99 & 21.65 \\
Rural/Dark (no streetlight) & 56.60 & 26.47 & 68.12 & 58.36 & 62.10 & 42.32 & 32.42 & 22.22 \\
OOD & 59.44 & 36.11 & 24.04 & 28.34 & 37.05 & 37.67 & 33.75 & 22.79 \\
\bottomrule
\end{tabular}
}
\end{table*}

Table~\ref{tab:scenario_performance} reports detection, classification, and system performance across four conditions: dense urban traffic and very high number of signboards, rain, rural roads without street lighting (Rural/Dark), and OOD sign appearances . All metrics are computed under streaming video inference on the Jetson Orin Nano. Please refer to Appendix \ref{sec:appen_qualitative_res} for visuals and failure cases.

\textbf{Finding 4: Detection and classification fail differently across conditions, revealing a differential bottleneck.} Under Rural/Dark conditions, detection quality degrades severely (mAP@50 56.60\%, mAP@50:95 26.47\%), but classification precision conditioned on successful detection remains competitive (A.P. 68.12\%). These results suggest that nighttime performance degradation is dominated primarily by the detection stage rather than the downstream classifier. The classifier operating on normalized, retroreflective-illuminated crops is relatively robust to absolute scene illumination, whereas the detector — which must locate signs against a dark, high-dynamic-range background — loses a substantial fraction of instances entirely. The implication for system improvement is specific: investment in nighttime detection (illumination-aware backbones~\cite{lin2025yolo, mishra2025benchmarking}, multi-exposure fusion) will yield larger system-level gains than equivalent investment in nighttime classification.

Under rain, mAP@50 drops to 61.20\% but classification recall improves slightly to 72.70\%: the hysteresis locking mechanism maintains previously stabilized labels during confidence suppression from precipitation, trading precision for recall stability. Throughput under rain is notably higher (38.46 FPS), a consequence of reduced sign density per frame lowering the classification invocation rate — a property of the decoupled architecture where computational load scales with detected object count, not frame rate.

\begin{table*}[t]
\centering
\caption[System-level performance of Edge-TSR under different frame sampling strategies]{
System-level performance under different frame sampling strategies on the Jetson Orin Nano, including FPS, latency in ms(Det., Cls., Trk.), CPU/GPU utilization (\%), and temperature ($^\circ C$).}
\label{tab:system_performance}
\resizebox{\textwidth}{!}{
\begin{tabular}{lc cc ccccc ccccc}
\toprule
\textbf{Setting} 
& \textbf{FPS} 
& \multicolumn{2}{c}{\textbf{Latency (ms)}} 
& \multicolumn{5}{c}{\textbf{CPU Statistics}} 
& \multicolumn{5}{c}{\textbf{GPU Statistics}} \\
\cmidrule(lr){3-4} \cmidrule(lr){5-9} \cmidrule(lr){10-14}
& 
& \textbf{Det + Trk} & \textbf{Cls} 
& \textbf{Util} & \textbf{Avg. Temp.} & \textbf{Max. Temp.} & \textbf{Avg. RAM} & \textbf{Max. RAM} 
& \textbf{Util} & \textbf{Avg. Temp.} & \textbf{Max. Temp.} & \textbf{Avg. Mem} & \textbf{Max. Mem} \\
\midrule

Every frame & 13.25 & 30.22 & 20.98 & 27.31 & 56.94 & 60.03 & 45.54 & 46.30 & 65.73 & 57.21 & 60.03 & 42.76 & 44.79 \\
Every $2^{nd}$ frame & 23.76 & 33.61 & 21.19 & 38.45 & 56.87 & 60.00 & 51.54 & 58.00 & 64.22 & 57.18 & 60.03 & 54.26 & 56.37 \\
Every $3^{rd}$ frame & 31.63 & 36.77 & 21.85 & 43.43 & 54.29 & 58.18 & 46.89 & 47.80 & 62.88 & 54.32 & 58.40 & 45.11 & 45.68 \\
Every $4^{th}$ frame & 41.12 & 37.41 & 21.98 & 53.60 & 56.06 & 58.97 & 50.85 & 51.70 & 64.17 & 56.14 & 58.94 & 54.36 & 55.96 \\
Every $5^{th}$ frame & 44.62 & 40.58 & 23.25 & 55.89 & 54.45 & 57.31 & 47.39 & 49.00 & 61.58 & 52.87 & 57.25 & 40.00 & 41.33 \\
Every $10^{th}$ frame & 62.72 & 50.01 & 27.77 & 68.32 & 51.41 & 54.56 & 46.89 & 47.00 & 59.08 & 50.87 & 54.25 & 41.82 & 42.85 \\
Every $30^{th}$ frame & 110.72 & 53.91 & 28.77 & 74.18 & 50.50 & 52.87 & 48.50 & 48.80 & 59.37 & 49.41 & 52.31 & 42.11 & 42.67 \\
\bottomrule
\end{tabular}
}
\end{table*}

\textbf{Finding 5: OOD performance reveals two distinct failure modes, predominantly localized to the classifier.} OOD classification metrics (A.P. 24.04\%, A.R. 28.34\%, accuracy 37.05\%) reflect two qualitatively different challenges: \textit{intra-class distribution shift} (known categories in non-standard physical forms, addressable via data augmentation) and \textit{true open-set failure} (categories absent from the 41-class vocabulary, requiring open-set recognition at inference time). Detection degrades more moderately (mAP@50 59.44\% vs.\ 73.12\% in Dense), indicating that the class-agnostic detector largely generalizes across appearance variation at the localization level, while the dominant penalty is borne by the closed-world classifier.

\textbf{Finding 6: Throughput is load-dependent, and the fastest conditions are the least accurate.} Throughput varies substantially across conditions (Dense: 31.63, Rain: 38.46, OOD: 37.67, Rural/Dark: 42.32 FPS). Since each classified region adds $\approx$20~ms (Section~\ref{sec:results}, RQ3), FPS tracks the number of regions classified per frame, which depends on both scene sign density and missed detections. The four videos differ markedly in sign density (71\%, 21\%, 43\%, and 13\% of frames contain signs in Dense, Rain, OOD, and Rural/Dark, respectively), and detector misses further reduce the classification load in adverse conditions. Consequently, across the four conditions, the fastest setting (Rural/Dark) is also the least accurate (mAP@50 56.60\%), and the slowest (Dense) the most accurate (73.12\%). System designers should therefore treat FPS as a load-dependent quantity rather than a fixed property of the hardware and model, and specify worst-case rather than best-case throughput.

\subsection{Sparse Sampling Trade-offs (RQ3)}

\begin{figure}[t]
    \centering
    
    \begin{subfigure}[t]{0.49\columnwidth}
        \centering
        \includegraphics[width=\linewidth]{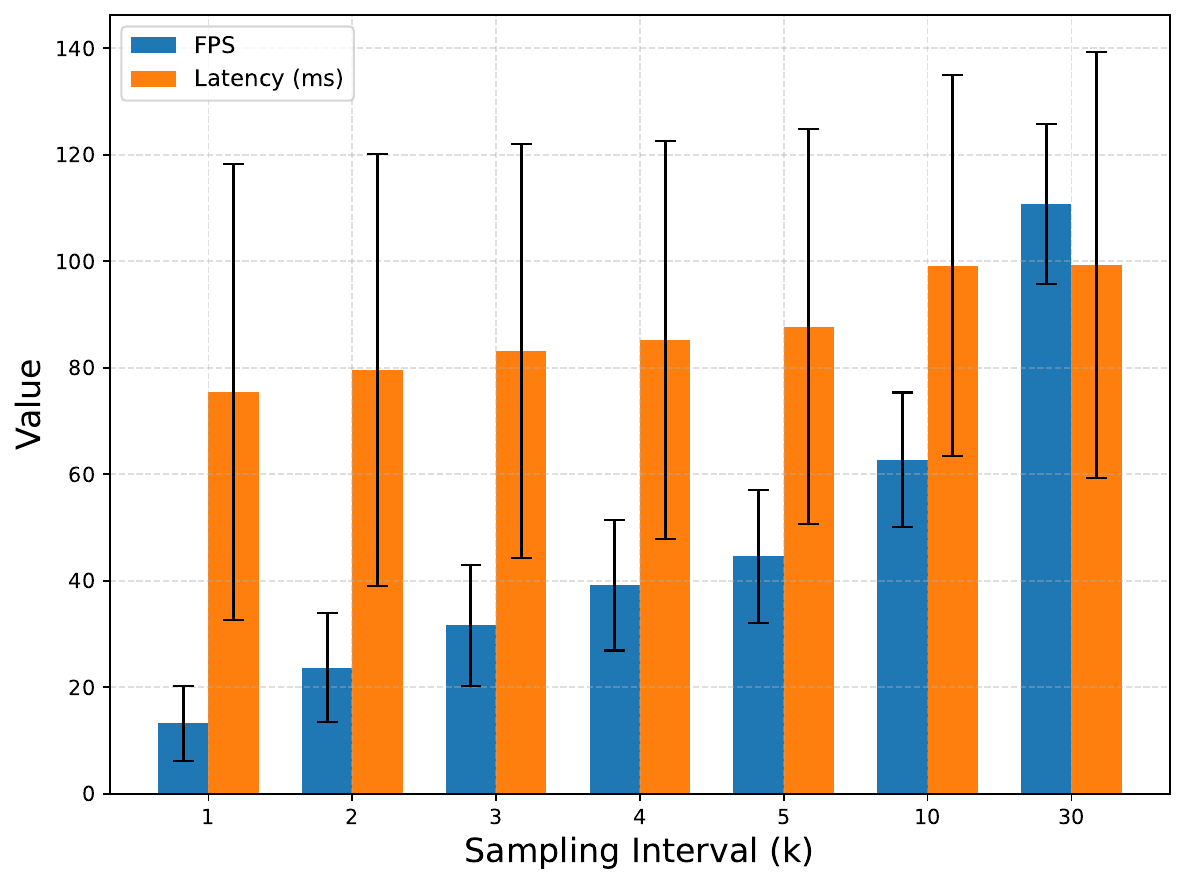}
        \caption{Inference stability}
        \label{fig:stability}
    \end{subfigure}
    \hfill
    \begin{subfigure}[t]{0.49\columnwidth}
        \centering
        \includegraphics[width=\linewidth]{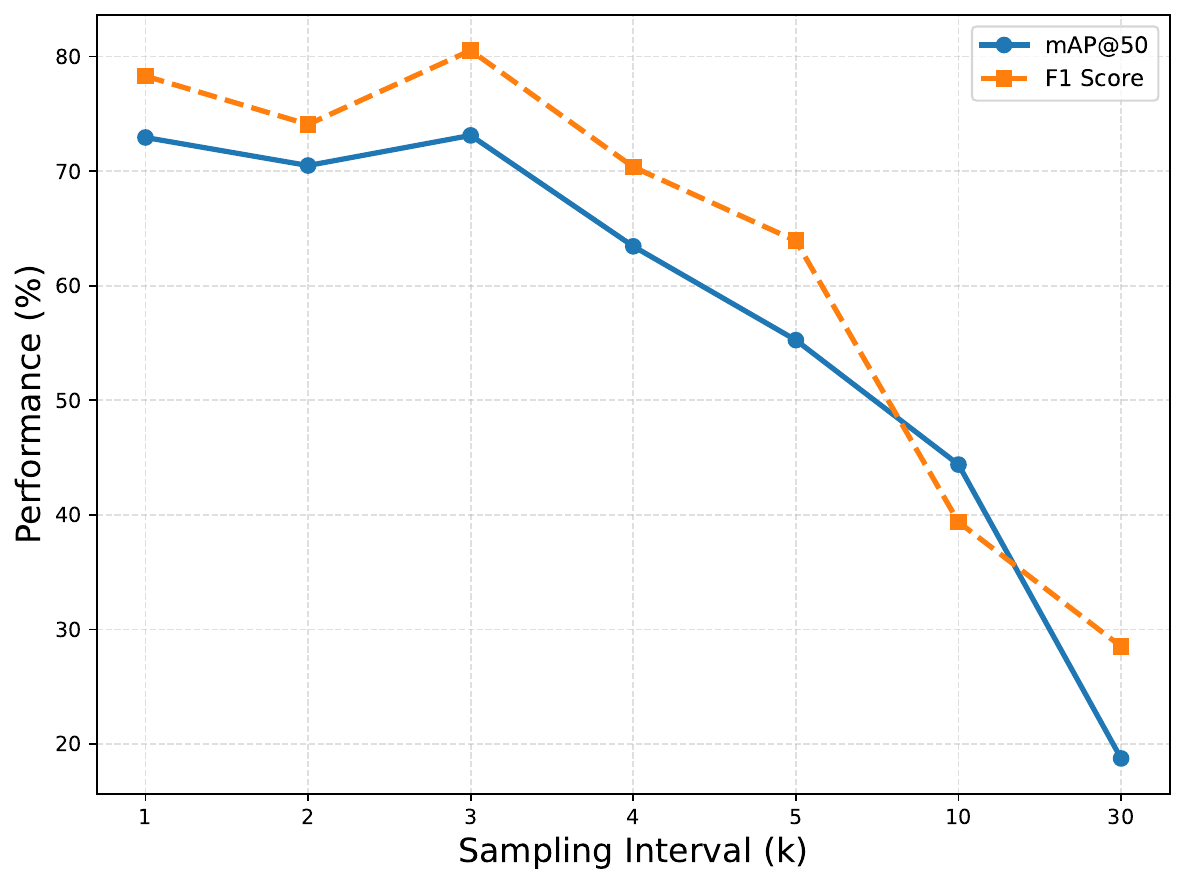}  
        \caption{Performance vs. sampling}
        \label{fig:map_f1}
    \end{subfigure}
    \caption[Effect of sparse frame sampling on system performance]{Effect of sparse frame sampling. Moderate sampling improves classification performance, larger intervals reduce performance and impact stability. Per-frame latency of sampled frames is plotted.}
    \label{fig:sampling_tradeoff}
\end{figure}

Tables~\ref{tab:system_performance} and Figure~\ref{fig:sampling_tradeoff}, characterize the effect of sampling interval $k$ on FPS, latency, thermal envelope, and recognition quality. Appendix~\ref{sec:appendix_latency} decomposes per-frame latency into sampled-frame and propagation-only costs for each $k$: the amortized latency matches $1000/\text{FPS}$ for every $k$, propagation costs only 4.8--6.7~ms per frame, and each classified region adds $\approx$20~ms.

\textbf{Finding 7: Every-frame inference falls below real-time.} At $k=1$, the system achieves only 13.25~FPS, below the 15~FPS threshold, while GPU utilization is 65.73\% and temperatures remain below 61\degree C. Every-frame inference is therefore limited by sequential per-region processing and host-side overhead rather than by thermal throttling or GPU saturation. \textbf{Adaptive sampling.} We also compare fixed $k=3$ with two runtime controllers that adapt $k \in [3,7]$ to measured load, optionally constrained by track-level accuracy signals (Appendix~\ref{sec:appendix_adaptive}). Both raise throughput by only 11--13\% while reducing recall by 19--26~pp, because load peaks coincide with sign-dense segments, where skipping observations is most costly. For this workload, a fixed interval at the accuracy optimum dominates load-adaptive sampling.

\textbf{Finding 8: $k=3$ is the optimal sampling interval.} To check whether reduced-precision inference changes the deployment picture, we also deployed the detector and classifier as TensorRT FP32 and FP16 engines at $k=3$ (Table ~\ref{tab:quantization}). TensorRT raised offline throughput only from 31 to 34~FPS, and FP16 gave no measurable gain over TensorRT FP32 (34.65 vs.\ 34.72~FPS). Detection performance is essentially unchanged (mAP@50 of 72.96 for both engines; classification accuracy within 0.1~pp), which indicates that end-to-end latency is dominated by host-side pre- and post-processing rather than network inference. Given this marginal throughput benefit, we report all main results with the unquantized PyTorch models, which keeps the pipeline simpler and fully reproducible. Increasing to $k{=}3$ yields a $2.38\times$ FPS improvement ($13.25 \rightarrow 31.63$) while simultaneously reducing average GPU temperature ($57.21 \rightarrow 54.32\degree$C) and CPU temperature ($56.94 \rightarrow 54.29\degree$C), as the reduced inference duty cycle lowers thermal accumulation below the threshold. Figure~\ref{fig:sampling_tradeoff}(b) shows that F1 score peaks at $k{=}3$ rather than $k{=}1$ — a non-monotonic relationship showing reduced inference frequency does not monotonically degrade recognition quality when the temporal stabilization module is active. Beyond $k{=}3$, both F1 and mAP decline monotonically as inference gaps grow large enough that tracker drift and temporal window starvation degrade stabilization quality.

Figure~\ref{fig:sampling_tradeoff}(a) shows throughput and latency variability across sampling intervals. The standard deviation of sampled-frame latency remains between 36 and 43~ms for all $k$, since it is driven by the number of regions per frame rather than by the sampling interval. Absolute FPS variability grows with $k$ (7 to 15~FPS), but relative variability decreases (coefficient of variation 0.53 at $k=1$, 0.36 at $k=3$, and 0.14 at $k=30$), as cheap non-sampled frames increasingly dominate each window.

Figure~\ref{fig:sampling_tradeoff}(b) shows that F1 peaks at $k=3$ (80.54 vs.\ 78.30 at $k=1$), while mAP@50 decreases only marginally from $k=1$ to $k=3$ (72.94 to 71.25). Reducing inference frequency therefore does not monotonically degrade recognition quality, and the temporal stabilization module aggregates sparse predictions into stable outputs up to $k=3$. Beyond $k=3$, both metrics decline monotonically (F1 28.51 and mAP@50 18.74 at $k=30$): with larger gaps, objects move further between sampled frames, so tracker association fails more often and fewer tracks are confirmed (1.92 to 1.35 classified regions per sampled frame from $k=1$ to $k=30$), while the per-track label window fills more slowly.

\textbf{Memory utilization is insensitive to sampling interval.} Table~\ref{tab:system_performance} shows that RAM and GPU memory utilization are largely invariant to $k$ (approximately 45--59\% RAM, 40--57\% GPU memory across all settings), with only minor fluctuations attributable to system-level variability.  This behavior is expected, as the memory footprint is dominated by model parameters and pre-allocated buffers, which remain constant regardless of inference frequency. Unlike compute utilization, which scales with the number of detector invocations, memory usage is largely independent of the sampling interval $k$. This indicates that sparse frame sampling can be applied at runtime without introducing additional memory pressure. While larger sampling intervals further increase throughput, recognition performance degrades substantially beyond $k=5$, indicating a clear trade-off between efficiency and accuracy.

\begin{table}[]
\centering
\caption[Ablation settings and their corresponding components of Edge-TSR]{Ablation configurations used to isolate the contribution of memory, voting, and hysteresis within the Edge-TSR temporal stabilization module.}
\label{tab:ablation_components}
\resizebox{0.95\columnwidth}{!}{
\begin{tabular}{lcccc}
\hline
\textbf{Condition} & \textbf{Memory} & \textbf{Voting} & \textbf{Hysteresis} & \textbf{What it isolates?} \\
\hline
No memory & $\times$ & $\times$ & $\times$ & Baseline: raw per-frame \\
No voting & $\checkmark$ & $\times$ & $\checkmark$ (conf-only) & Value of count-based voting \\
No hysteresis & $\checkmark$ & $\checkmark$ & $\times$ & Value of locking \\
Edge-TSR & $\checkmark$ & $\checkmark$ & $\checkmark$ & Complete system \\
\hline
\end{tabular}
}
\end{table}

\begin{figure}[t]
    \centering
    \includegraphics[width=0.6\columnwidth]{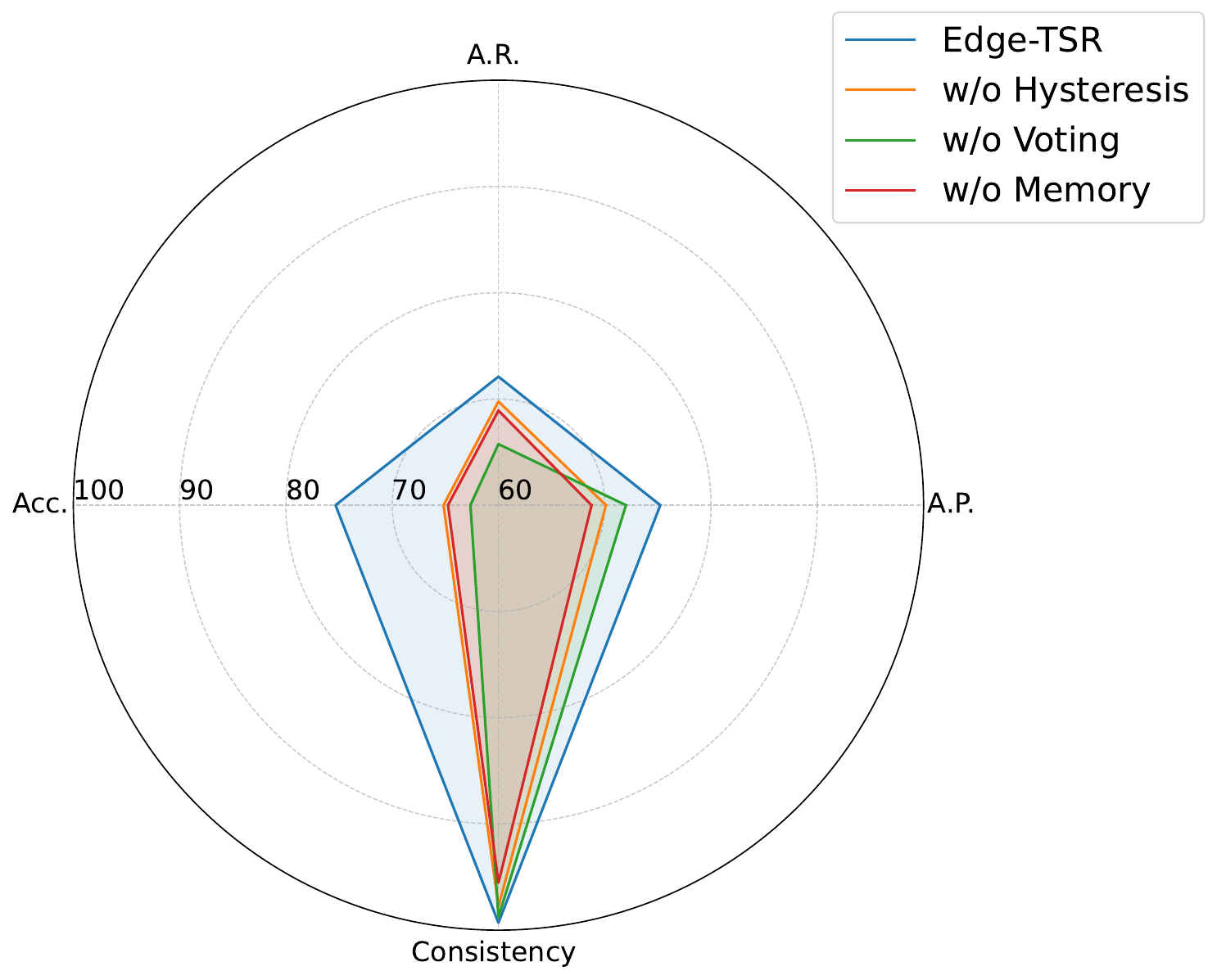}
    \caption[Ablation study of Edge-TSR]{Ablation of temporal components in the Edge-TSR.}
    \label{fig:ablation-edge}
    \vspace{-1.5em}
\end{figure}

\subsection{Ablation Study} \label{sec:ablation}
Figure~\ref{fig:ablation-edge} reports A.P., A.R., accuracy, and label consistency across four configurations defined in Table~\ref{tab:ablation_components}. 

\textbf{Finding 9: Hysteresis locking suppresses label oscillations under sparse sampling.} Removing hysteresis reduces accuracy from 75.33\% to 65.17\% (-10.16~pp) and increases the label flip rate from 0.69\% to 2.23\%, more than tripling label oscillations. Voting alone therefore cannot prevent accuracy-degrading oscillations, because per-frame classifier noise under sparse sampling is substantial. Hysteresis absorbs these transient misclassifications by requiring sustained, high-confidence contradictory evidence before a locked label is changed.

\textbf{Finding 10: Count-based voting is necessary for accurate locking.} Replacing count-based voting with confidence-only locking (\textit{No Voting}) causes the largest accuracy drop among the ablated components, from 75.33\% to 62.64\% (-12.69~pp). Compared with \textit{No Hysteresis}, this variant achieves higher consistency (98.76\% vs.\ 97.77\%) but lower accuracy (62.64\% vs.\ 65.17\%) and recall (65.75\% vs.\ 69.76\%, -4.01~pp). Confidence-only locking commits on the first high-confidence prediction: this suppresses flips effectively, but it also locks transient misclassifications and blocks legitimate label changes as signs enter or leave the frame. Count-based voting instead requires a label to appear in at least 3 of the 5 most recent observations before locking, which filters noise while remaining responsive to genuine sign transitions.

\textbf{Finding 11: The full system's gains are synergistic, not additive.} Neither component alone improves accuracy over \textit{No Memory} (64.75\%): \textit{No Hysteresis} reaches 65.17\% (+0.42~pp) and \textit{No Voting} 62.64\% (-2.11~pp), yet combined they yield 75.33\% (+10.58~pp), confirming that the gains are synergistic: count-based voting produces a higher-quality candidate label that hysteresis then stabilizes.

\subsection{Real-World Vehicular Deployment (RQ4)}
\begin{figure}[t]
    \centering
    \includegraphics[width=0.9\columnwidth]{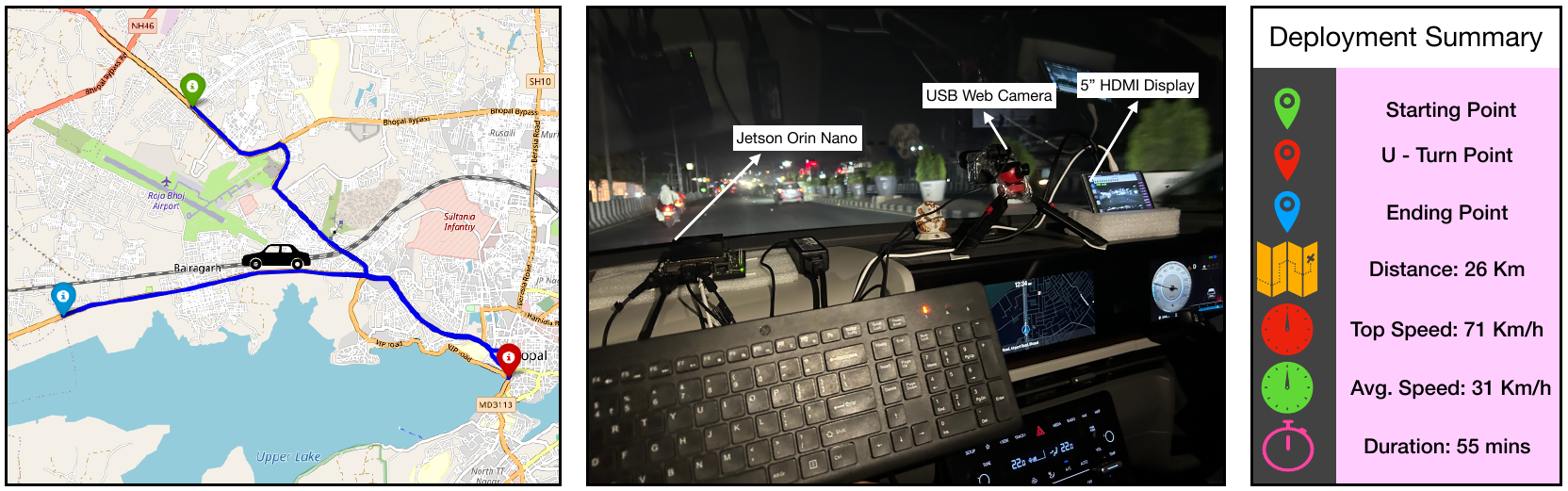}
    \caption[Real-world deployment summary]{
    Real-world deployment summary showing the traversed route and key motion statistics, including distance, duration, average speed, and maximum speed.}
    \label{fig:deployment-summary}
\end{figure}

\begin{figure}[t]
    \centering
    \includegraphics[width=0.9\columnwidth]{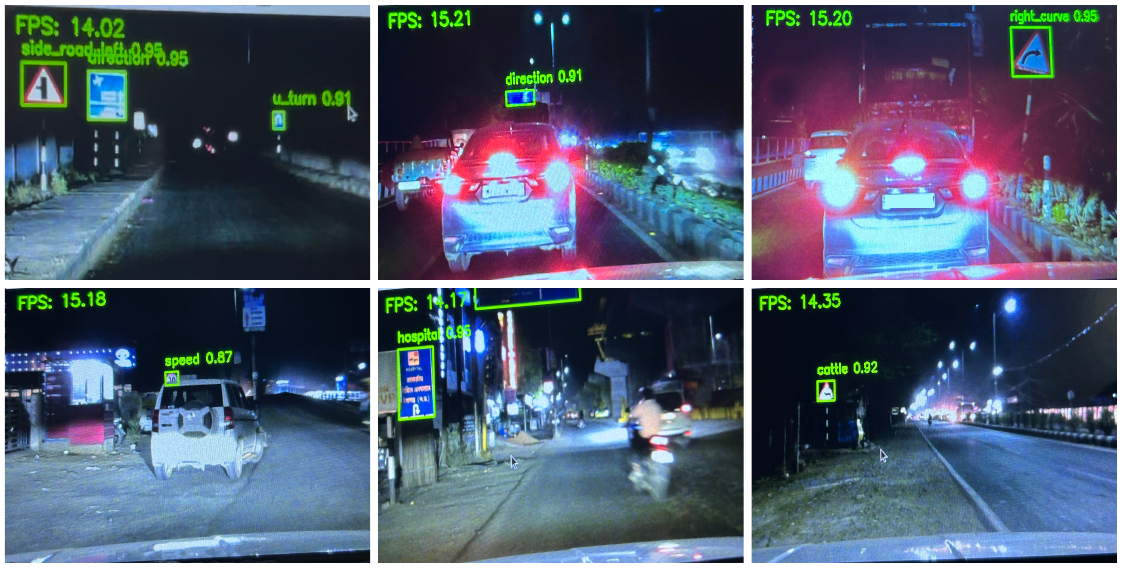}
    \caption[Qualitative results from real-world deployment of Edge-TSR]{Qualitative results from real-world deployment of Edge-TSR, showing  detection and classification of diverse signs under unconstrained driving conditions.}
    \label{fig:real-world-predictions}
\end{figure}

We conducted a continuous 55-minute deployment trial over a 26~km urban and peri-urban route (Figure~\ref{fig:deployment-summary}), with average speed 31~km/h and peak speed 71~km/h. The system made 10,371 recognition decisions over the trial duration. Figure~\ref{fig:real-world-predictions} shows representative frames illustrating correct recognition of diverse sign categories under nighttime conditions with mixed artificial illumination.

\textbf{Finding 12: Component latency increases under live deployment.} The system achieved a mean online throughput of 16.18~FPS, above the 15~FPS real-time threshold with 7.8\% headroom~\cite{wagner2009real}. Per-invocation latency, however, was substantially higher than in controlled evaluation: Det+Trk increased from 36.77 to 65.68~ms and classification from 21.85 to 34.97~ms per region (+79\% and +60\%). We attribute this increase to system-level effects present only in live operation, likely including camera capture, concurrent logging, and clock scaling; we did not isolate their individual contributions. The reduction from 31.63~FPS offline to 16.18~FPS online is consistent with these larger per-invocation latencies, together with per-frame capture and display costs that are absent from headless offline evaluation. The practical implication is that component-level latency benchmarks are lower bounds on deployed latency, not deployment targets, and that headroom must be reserved explicitly for system overhead.

\textbf{Finding 13: The thermal envelope remains safe throughout, with vehicular airflow providing passive cooling benefit.} Average GPU temperature during the trial was 48.15$\degree$C (max 50.37$\degree$C) and average CPU temperature was 48.78$\degree$C (max 50.68$\degree$C) — both substantially below the controlled evaluation figures (GPU avg. 54.32$\degree$C at $k{=}3$). This reduction is attributable to passive airflow from the vehicle's air conditioning system reducing ambient temperature around the device, confirming that the thermal management concerns identified under static conditions are not a bottleneck in real vehicular deployment.

\textbf{At-speed recognition boundary.} At peak speed 71~km/h (${\approx}20$~m/s) with effective detection rate ${\approx}5$~Hz under $k{=}3$, the vehicle travels ${\approx}4$~m between detection frames. For signs visible for 2--3 seconds (${\approx}40$--60~m at 71~km/h), the temporal window $T{=}5$ accumulates sufficient observations to lock a label before the sign exits the field of view. For signs with shorter visibility windows — junction advisories, highway exit ramps — there is a risk of insufficient temporal accumulation before the sign leaves frame, a failure mode observed in a small number of high-speed frames during the trial and not recoverable by the temporal module without a higher detection rate or speed-adaptive sampling.
 
\section{Discussion}
\label{sec:discussion}

\textbf{\textit{Deployment Reveals What Benchmarks Cannot.}} The image-to-video performance gap documented in Section~\ref{sec:results} — 20--30~pp in detection mAP and up to 20.90~pp in classification across three independent baseline systems — is not an artifact of any particular architecture or dataset. The observed gap appears to arise from evaluating systems on independent image samples drawn from a stationary distribution while deployment occurs over temporally correlated, non-stationary video streams subject to motion-induced degradation. This gap is large enough to reverse the performance ordering of systems: LENS-Net, the strongest image-based classifier, falls to third among baselines under video evaluation. Our results suggest that TSR systems intended for reliable field deployment should be evaluated under continuous streaming conditions in addition to conventional image benchmarks. This methodological finding is not unique to traffic signs — any embedded AI pipeline whose inputs are drawn from a continuous, environment-coupled sensor stream will exhibit analogous degradation when per-frame models are deployed without explicit temporal reasoning~\cite{xu2022mandheling, bianco2018benchmark}. 

\textbf{\textit{Temporal Reasoning as Systems Infrastructure.}} The ablation study establishes that temporal stability is a first-class system requirement, not an optional post-processing refinement. Removing hysteresis locking alone degrades accuracy by a margin (10.16~pp) larger than the accuracy gap between the strongest and weakest image-based baselines in Table~\ref{table:results}. In a deployment context where a label oscillation on a speed-limit or right-of-way sign produces a conflicting constraint that a downstream planning module must resolve, label flickering is not a marginal quality-of-life issue — it is a distinct failure mode with direct operational consequences. The $O(T)$ per-track cost of the full stabilization module at $T{=}5$ is negligible relative to detection and classification, making the cost-benefit ratio of temporal stabilization highly favorable for any resource-constrained deployment.

\textbf{\textit{Design Lessons for Continuous Edge Inference.}} Three findings from this work are tied to system architecture rather than to TSR itself, and are therefore likely to transfer to other continuous edge inference pipelines, although we evaluate them only for TSR on the Jetson Orin Nano.

\noindent \textit{$\bullet$ Load-proportional cost creates a throughput-accuracy inversion:} In decoupled detection-classification architectures, computational demand scales with detected object count rather than frame resolution. Conditions that suppress the detector — adverse weather, sparse scenes — increase throughput while simultaneously degrading recognition quality. System designers must treat worst-case FPS as a load-dependent quantity rather than a fixed hardware-model property, and should not use best-case throughput figures in deployment specifications. 

\noindent \textit{$\bullet$ Component benchmarks are upper bounds, not deployment targets:} Detector latency approximately doubles under live vehicular operation relative to controlled evaluation (36.77~ms $\rightarrow$ 65.68~ms), dominated by system-level overhead rather than model computation. This gap is invisible in component-level benchmarks and can only be characterized through end-to-end deployment trials. Embedded system specifications should reserve headroom for OS scheduling, display rendering, and concurrent process contention. 

\noindent \textit{$\bullet$ Thermal management is context-dependent:} GPU temperatures during live deployment were about 6$\degree$C lower than in controlled evaluation, attributable to passive vehicular airflow. This favorable interaction cannot be relied upon universally — enclosed vehicles, warm climates, or slow-moving traffic may eliminate this benefit — but it illustrates that real deployment contexts introduce environmental variables that static bench testing cannot capture. Thermal-resilient system design should be considered standard practice for any continuous-inference embedded AI system.

\noindent \textit{Adapting Edge-TSR.} The pipeline applies without changes to other detect-then-classify tasks with persistent objects (e.g., traffic-light state or license-plate recognition), requiring only retrained models and re-tuned stabilization thresholds. On other hardware, the sampling interval follows directly from the latency model of Section~\ref{sec:results}: measuring $L_s$ and $L_p$ on the target platform,    the smallest interval meeting a per-frame time budget $B$ (ms) is $k_{\min} = \lceil (L_s - L_p)/(B - L_p) \rceil$, valid for $B > L_p$.

\textbf{\textit{Limitations and Future Directions.}} Several limitations of the current system motivate important directions for future work.
\textbf{(i) Closed-world classification.} The classifier operates under a fixed 41-class vocabulary and therefore cannot reliably abstain on unseen traffic sign categories. The OOD evaluation demonstrates that the dominant failure mode is high-confidence misclassification of semantically unknown signs rather than localization failure. Incorporating open-set recognition mechanisms, such as energy-based or prototype-distance novelty detection \cite{liu2020energy}, would enable the system to reject unfamiliar inputs instead of producing confidently incorrect predictions.
\textbf{(ii) Geographic and deployment scope.} Edge-TSR is trained and evaluated primarily on Indian road environments using INTSD \cite{uikey2024indian, mishra2025benchmarking}, and the deployment evaluation dataset was collected within a single city. While the deployment-centric evaluation methodology generalizes beyond this setting, quantitative performance figures should not be assumed to transfer directly across countries, roadway structures, or traffic environments. Broader deployment trials spanning diverse geographic regions, vehicle types, camera configurations, and sustained highway-speed operation would strengthen the generalizability of the findings.
\textbf{(iii) Single-modality sensing.} The current system relies monocular RGB sensing and therefore remains vulnerable to illumination degradation, adverse weather, and other photometric failure modes. Integrating complementary sensing modalities such as infrared imaging, radar, or depth sensing \cite{bijelic2020seeing} may improve deployment robustness under challenging environmental conditions, albeit at increased system complexity, power draw, and deployment cost.

\section{Conclusion}

 We present Edge-TSR, a deployment-oriented continuous edge inference system for traffic sign recognition on the Jetson Orin Nano, and characterize its behavior under real-world streaming conditions that image-based evaluation protocols do not capture. Evaluating on the source videos of INTSD, we showed that state-of-the-art TSR models lose 20--30 percentage points of mAP@50 when moving from image-based to streaming evaluation, indicating that image benchmarks systematically overstate deployed TSR performance. On streaming video, Edge-TSR achieves the best classification performance among all evaluated baselines (75.23\% macro-averaged precision on the dense urban video), and lightweight temporal stabilization improves accuracy by 10.58 percentage points over per-frame inference at 0.07~ms per sampled frame. A 55-minute vehicular deployment demonstrates sustained online operation at 16.18~FPS within safe thermal limits. Beyond recognition quality, our characterization reveals system-level behaviors that controlled benchmarking does not expose: load-proportional throughput, latency inflation under live operation, and context-dependent thermal behavior. These results suggest that streaming, deployment-aware evaluation and stateful temporal inference should be first-class design considerations for embedded TSR. We release our annotated 50,732-frame streaming evaluation dataset and full system implementation to support reproducible deployment-centric evaluation.

\bibliographystyle{acm-reference-format}
\bibliography{ref}


\begin{thebibliography}{53}


\ifx \showCODEN    \undefined \def \showCODEN     #1{\unskip}     \fi
\ifx \showISBNx    \undefined \def \showISBNx     #1{\unskip}     \fi
\ifx \showISBNxiii \undefined \def \showISBNxiii  #1{\unskip}     \fi
\ifx \showISSN     \undefined \def \showISSN      #1{\unskip}     \fi
\ifx \showLCCN     \undefined \def \showLCCN      #1{\unskip}     \fi
\ifx \shownote     \undefined \def \shownote      #1{#1}          \fi
\ifx \showarticletitle \undefined \def \showarticletitle #1{#1}   \fi
\ifx \showURL      \undefined \def \showURL       {\relax}        \fi
\providecommand\bibfield[2]{#2}
\providecommand\bibinfo[2]{#2}
\providecommand\natexlab[1]{#1}
\providecommand\showeprint[2][]{arXiv:#2}

\bibitem[Aharon et~al\mbox{.}(2022)]%
        {aharon2022bot}
\bibfield{author}{\bibinfo{person}{Nir Aharon}, \bibinfo{person}{Roy Orfaig}, {and} \bibinfo{person}{Ben-Zion Bobrovsky}.} \bibinfo{year}{2022}\natexlab{}.
\newblock \showarticletitle{BoT-SORT: Robust associations multi-pedestrian tracking}.
\newblock \bibinfo{journal}{\emph{arXiv preprint arXiv:2206.14651}} (\bibinfo{year}{2022}).
\newblock


\bibitem[Anoop et~al\mbox{.}(2025)]%
        {anoop2025real}
\bibfield{author}{\bibinfo{person}{KS Anoop}, \bibinfo{person}{KK Chandrathejas}, \bibinfo{person}{SP Anusha}, {et~al\mbox{.}}} \bibinfo{year}{2025}\natexlab{}.
\newblock \showarticletitle{Real-Time Two-Stage Detection of Indian Traffic Signboards Using YOLO11 on Jetson Orin Nano}. In \bibinfo{booktitle}{\emph{2025 International Conference on Advancements in Power, Communication and Intelligent Systems (APCI)}}. IEEE, \bibinfo{pages}{1--6}.
\newblock


\bibitem[Ayachi et~al\mbox{.}(2022)]%
        {ayachi2022edge}
\bibfield{author}{\bibinfo{person}{Riadh Ayachi}, \bibinfo{person}{Mouna Afif}, \bibinfo{person}{Yahia Said}, {and} \bibinfo{person}{Abdessalem Ben~Abdelali}.} \bibinfo{year}{2022}\natexlab{}.
\newblock \showarticletitle{An edge implementation of a traffic sign detection system for advanced driver assistance systems}.
\newblock \bibinfo{journal}{\emph{International Journal of Intelligent Robotics and Applications}} \bibinfo{volume}{6}, \bibinfo{number}{2} (\bibinfo{year}{2022}), \bibinfo{pages}{207--215}.
\newblock


\bibitem[Benoit-Cattin et~al\mbox{.}(2020)]%
        {benoit2020impact}
\bibfield{author}{\bibinfo{person}{Th{\'e}o Benoit-Cattin}, \bibinfo{person}{Delia Velasco-Montero}, {and} \bibinfo{person}{Jorge Fern{\'a}ndez-Berni}.} \bibinfo{year}{2020}\natexlab{}.
\newblock \showarticletitle{Impact of thermal throttling on long-term visual inference in a CPU-based edge device}.
\newblock \bibinfo{journal}{\emph{Electronics}} \bibinfo{volume}{9}, \bibinfo{number}{12} (\bibinfo{year}{2020}), \bibinfo{pages}{2106}.
\newblock


\bibitem[Bewley et~al\mbox{.}(2016)]%
        {bewley2016simple}
\bibfield{author}{\bibinfo{person}{Alex Bewley}, \bibinfo{person}{Zongyuan Ge}, \bibinfo{person}{Lionel Ott}, \bibinfo{person}{Fabio Ramos}, {and} \bibinfo{person}{Ben Upcroft}.} \bibinfo{year}{2016}\natexlab{}.
\newblock \showarticletitle{Simple online and realtime tracking}. In \bibinfo{booktitle}{\emph{2016 IEEE international conference on image processing (ICIP)}}. Ieee, \bibinfo{pages}{3464--3468}.
\newblock


\bibitem[Bhardwaj et~al\mbox{.}(2022)]%
        {bhardwaj2022ekya}
\bibfield{author}{\bibinfo{person}{Romil Bhardwaj}, \bibinfo{person}{Zhengxu Xia}, \bibinfo{person}{Ganesh Ananthanarayanan}, \bibinfo{person}{Junchen Jiang}, \bibinfo{person}{Yuanchao Shu}, \bibinfo{person}{Nikolaos Karianakis}, \bibinfo{person}{Kevin Hsieh}, \bibinfo{person}{Paramvir Bahl}, {and} \bibinfo{person}{Ion Stoica}.} \bibinfo{year}{2022}\natexlab{}.
\newblock \showarticletitle{Ekya: Continuous learning of video analytics models on edge compute servers}. In \bibinfo{booktitle}{\emph{19th USENIX Symposium on Networked Systems Design and Implementation (NSDI 22)}}. \bibinfo{pages}{119--135}.
\newblock


\bibitem[Bianco et~al\mbox{.}(2018)]%
        {bianco2018benchmark}
\bibfield{author}{\bibinfo{person}{Simone Bianco}, \bibinfo{person}{Remi Cadene}, \bibinfo{person}{Luigi Celona}, {and} \bibinfo{person}{Paolo Napoletano}.} \bibinfo{year}{2018}\natexlab{}.
\newblock \showarticletitle{Benchmark analysis of representative deep neural network architectures}.
\newblock \bibinfo{journal}{\emph{IEEE access}}  \bibinfo{volume}{6} (\bibinfo{year}{2018}), \bibinfo{pages}{64270--64277}.
\newblock


\bibitem[Bijelic et~al\mbox{.}(2020)]%
        {bijelic2020seeing}
\bibfield{author}{\bibinfo{person}{Mario Bijelic}, \bibinfo{person}{Tobias Gruber}, \bibinfo{person}{Fahim Mannan}, \bibinfo{person}{Florian Kraus}, \bibinfo{person}{Werner Ritter}, \bibinfo{person}{Klaus Dietmayer}, {and} \bibinfo{person}{Felix Heide}.} \bibinfo{year}{2020}\natexlab{}.
\newblock \showarticletitle{Seeing through fog without seeing fog: Deep multimodal sensor fusion in unseen adverse weather}. In \bibinfo{booktitle}{\emph{Proceedings of the IEEE/CVF conference on computer vision and pattern recognition}}. \bibinfo{pages}{11682--11692}.
\newblock


\bibitem[Canziani et~al\mbox{.}(2016)]%
        {canziani2016analysis}
\bibfield{author}{\bibinfo{person}{Alfredo Canziani}, \bibinfo{person}{Adam Paszke}, {and} \bibinfo{person}{Eugenio Culurciello}.} \bibinfo{year}{2016}\natexlab{}.
\newblock \showarticletitle{An analysis of deep neural network models for practical applications}.
\newblock \bibinfo{journal}{\emph{arXiv preprint arXiv:1605.07678}} (\bibinfo{year}{2016}).
\newblock


\bibitem[Chen et~al\mbox{.}(2025)]%
        {chen2025yolo}
\bibfield{author}{\bibinfo{person}{Junzhou Chen}, \bibinfo{person}{Heqiang Huang}, \bibinfo{person}{Ronghui Zhang}, \bibinfo{person}{Nengchao Lyu}, \bibinfo{person}{Yanyong Guo}, \bibinfo{person}{Hong-Ning Dai}, {and} \bibinfo{person}{Hong Yan}.} \bibinfo{year}{2025}\natexlab{}.
\newblock \showarticletitle{Yolo-ts: Real-time traffic sign detection with enhanced accuracy using optimized receptive fields and anchor-free fusion}.
\newblock \bibinfo{journal}{\emph{IEEE Transactions on Intelligent Transportation Systems}} (\bibinfo{year}{2025}).
\newblock


\bibitem[{CVAT.ai Corporation}(2023)]%
        {cvat}
\bibfield{author}{\bibinfo{person}{{CVAT.ai Corporation}}.} \bibinfo{year}{2023}\natexlab{}.
\newblock \bibinfo{booktitle}{\emph{Computer Vision Annotation Tool (CVAT)}}.
\newblock
\href{https://doi.org/10.5281/zenodo.4009388}{doi:\nolinkurl{10.5281/zenodo.4009388}}


\bibitem[Du et~al\mbox{.}(2023)]%
        {du2023strongsort}
\bibfield{author}{\bibinfo{person}{Yunhao Du}, \bibinfo{person}{Zhicheng Zhao}, \bibinfo{person}{Yang Song}, \bibinfo{person}{Yanyun Zhao}, \bibinfo{person}{Fei Su}, \bibinfo{person}{Tao Gong}, {and} \bibinfo{person}{Hongying Meng}.} \bibinfo{year}{2023}\natexlab{}.
\newblock \showarticletitle{Strongsort: Make deepsort great again}.
\newblock \bibinfo{journal}{\emph{IEEE Transactions on Multimedia}}  \bibinfo{volume}{25} (\bibinfo{year}{2023}), \bibinfo{pages}{8725--8737}.
\newblock


\bibitem[Ertler et~al\mbox{.}(2020)]%
        {ertler2020mapillary}
\bibfield{author}{\bibinfo{person}{Christian Ertler}, \bibinfo{person}{Jerneja Mislej}, \bibinfo{person}{Tobias Ollmann}, \bibinfo{person}{Lorenzo Porzi}, \bibinfo{person}{Gerhard Neuhold}, {and} \bibinfo{person}{Yubin Kuang}.} \bibinfo{year}{2020}\natexlab{}.
\newblock \showarticletitle{The mapillary traffic sign dataset for detection and classification on a global scale}. In \bibinfo{booktitle}{\emph{European conference on computer vision}}. Springer, \bibinfo{pages}{68--84}.
\newblock


\bibitem[Everingham et~al\mbox{.}(2010)]%
        {everingham2010pascal}
\bibfield{author}{\bibinfo{person}{Mark Everingham}, \bibinfo{person}{Luc Van~Gool}, \bibinfo{person}{Christopher~KI Williams}, \bibinfo{person}{John Winn}, {and} \bibinfo{person}{Andrew Zisserman}.} \bibinfo{year}{2010}\natexlab{}.
\newblock \showarticletitle{The pascal visual object classes (voc) challenge}.
\newblock \bibinfo{journal}{\emph{International journal of computer vision}} \bibinfo{volume}{88}, \bibinfo{number}{2} (\bibinfo{year}{2010}), \bibinfo{pages}{303--338}.
\newblock


\bibitem[Eykholt et~al\mbox{.}(2018)]%
        {eykholt2018robust}
\bibfield{author}{\bibinfo{person}{Kevin Eykholt}, \bibinfo{person}{Ivan Evtimov}, \bibinfo{person}{Earlence Fernandes}, \bibinfo{person}{Bo Li}, \bibinfo{person}{Amir Rahmati}, \bibinfo{person}{Chaowei Xiao}, \bibinfo{person}{Atul Prakash}, \bibinfo{person}{Tadayoshi Kohno}, {and} \bibinfo{person}{Dawn Song}.} \bibinfo{year}{2018}\natexlab{}.
\newblock \showarticletitle{Robust physical-world attacks on deep learning visual classification}. In \bibinfo{booktitle}{\emph{Proceedings of the IEEE conference on computer vision and pattern recognition}}. \bibinfo{pages}{1625--1634}.
\newblock


\bibitem[Fang et~al\mbox{.}(2018)]%
        {fang2018nestdnn}
\bibfield{author}{\bibinfo{person}{Biyi Fang}, \bibinfo{person}{Xiao Zeng}, {and} \bibinfo{person}{Mi Zhang}.} \bibinfo{year}{2018}\natexlab{}.
\newblock \showarticletitle{Nestdnn: Resource-aware multi-tenant on-device deep learning for continuous mobile vision}. In \bibinfo{booktitle}{\emph{Proceedings of the 24th Annual International Conference on Mobile Computing and Networking}}. \bibinfo{pages}{115--127}.
\newblock


\bibitem[Han et~al\mbox{.}(2020)]%
        {han2020mining}
\bibfield{author}{\bibinfo{person}{Mingfei Han}, \bibinfo{person}{Yali Wang}, \bibinfo{person}{Xiaojun Chang}, {and} \bibinfo{person}{Yu Qiao}.} \bibinfo{year}{2020}\natexlab{}.
\newblock \showarticletitle{Mining inter-video proposal relations for video object detection}. In \bibinfo{booktitle}{\emph{European conference on computer vision}}. Springer, \bibinfo{pages}{431--446}.
\newblock


\bibitem[Han et~al\mbox{.}(2015)]%
        {han2015deep}
\bibfield{author}{\bibinfo{person}{Song Han}, \bibinfo{person}{Huizi Mao}, {and} \bibinfo{person}{William~J Dally}.} \bibinfo{year}{2015}\natexlab{}.
\newblock \showarticletitle{Deep compression: Compressing deep neural networks with pruning, trained quantization and huffman coding}.
\newblock \bibinfo{journal}{\emph{arXiv preprint arXiv:1510.00149}} (\bibinfo{year}{2015}).
\newblock


\bibitem[He et~al\mbox{.}(2016)]%
        {he2016deep}
\bibfield{author}{\bibinfo{person}{Kaiming He}, \bibinfo{person}{Xiangyu Zhang}, \bibinfo{person}{Shaoqing Ren}, {and} \bibinfo{person}{Jian Sun}.} \bibinfo{year}{2016}\natexlab{}.
\newblock \showarticletitle{Deep residual learning for image recognition}. In \bibinfo{booktitle}{\emph{Proceedings of the IEEE conference on computer vision and pattern recognition}}. \bibinfo{pages}{770--778}.
\newblock


\bibitem[Hinton et~al\mbox{.}(2015)]%
        {hinton2015distilling}
\bibfield{author}{\bibinfo{person}{Geoffrey Hinton}, \bibinfo{person}{Oriol Vinyals}, {and} \bibinfo{person}{Jeff Dean}.} \bibinfo{year}{2015}\natexlab{}.
\newblock \showarticletitle{Distilling the knowledge in a neural network}.
\newblock \bibinfo{journal}{\emph{arXiv preprint arXiv:1503.02531}} (\bibinfo{year}{2015}).
\newblock


\bibitem[Howard et~al\mbox{.}(2019)]%
        {howard2019searching}
\bibfield{author}{\bibinfo{person}{Andrew Howard}, \bibinfo{person}{Mark Sandler}, \bibinfo{person}{Grace Chu}, \bibinfo{person}{Liang-Chieh Chen}, \bibinfo{person}{Bo Chen}, \bibinfo{person}{Mingxing Tan}, \bibinfo{person}{Weijun Wang}, \bibinfo{person}{Yukun Zhu}, \bibinfo{person}{Ruoming Pang}, \bibinfo{person}{Vijay Vasudevan}, {et~al\mbox{.}}} \bibinfo{year}{2019}\natexlab{}.
\newblock \showarticletitle{Searching for mobilenetv3}. In \bibinfo{booktitle}{\emph{Proceedings of the IEEE/CVF international conference on computer vision}}. \bibinfo{pages}{1314--1324}.
\newblock


\bibitem[Jacob et~al\mbox{.}(2018)]%
        {jacob2018quantization}
\bibfield{author}{\bibinfo{person}{Benoit Jacob}, \bibinfo{person}{Skirmantas Kligys}, \bibinfo{person}{Bo Chen}, \bibinfo{person}{Menglong Zhu}, \bibinfo{person}{Matthew Tang}, \bibinfo{person}{Andrew Howard}, \bibinfo{person}{Hartwig Adam}, {and} \bibinfo{person}{Dmitry Kalenichenko}.} \bibinfo{year}{2018}\natexlab{}.
\newblock \showarticletitle{Quantization and training of neural networks for efficient integer-arithmetic-only inference}. In \bibinfo{booktitle}{\emph{Proceedings of the IEEE conference on computer vision and pattern recognition}}. \bibinfo{pages}{2704--2713}.
\newblock


\bibitem[Jiang et~al\mbox{.}(2018)]%
        {jiang2018chameleon}
\bibfield{author}{\bibinfo{person}{Junchen Jiang}, \bibinfo{person}{Ganesh Ananthanarayanan}, \bibinfo{person}{Peter Bodik}, \bibinfo{person}{Siddhartha Sen}, {and} \bibinfo{person}{Ion Stoica}.} \bibinfo{year}{2018}\natexlab{}.
\newblock \showarticletitle{Chameleon: scalable adaptation of video analytics}. In \bibinfo{booktitle}{\emph{Proceedings of the 2018 conference of the ACM special interest group on data communication}}. \bibinfo{pages}{253--266}.
\newblock


\bibitem[Jocher et~al\mbox{.}(2023)]%
        {yolov8_ultralytics}
\bibfield{author}{\bibinfo{person}{Glenn Jocher}, \bibinfo{person}{Ayush Chaurasia}, {and} \bibinfo{person}{Jing Qiu}.} \bibinfo{year}{2023}\natexlab{}.
\newblock \bibinfo{title}{Ultralytics YOLOv8}.
\newblock
\urldef\tempurl%
\url{https://github.com/ultralytics/ultralytics}
\showURL{%
\tempurl}


\bibitem[Koh et~al\mbox{.}(2021)]%
        {koh2021wilds}
\bibfield{author}{\bibinfo{person}{Pang~Wei Koh}, \bibinfo{person}{Shiori Sagawa}, \bibinfo{person}{Henrik Marklund}, \bibinfo{person}{Sang~Michael Xie}, \bibinfo{person}{Marvin Zhang}, \bibinfo{person}{Akshay Balsubramani}, \bibinfo{person}{Weihua Hu}, \bibinfo{person}{Michihiro Yasunaga}, \bibinfo{person}{Richard~Lanas Phillips}, \bibinfo{person}{Irena Gao}, {et~al\mbox{.}}} \bibinfo{year}{2021}\natexlab{}.
\newblock \showarticletitle{Wilds: A benchmark of in-the-wild distribution shifts}. In \bibinfo{booktitle}{\emph{International conference on machine learning}}. PMLR, \bibinfo{pages}{5637--5664}.
\newblock


\bibitem[Li et~al\mbox{.}(2017)]%
        {li2017perceptual}
\bibfield{author}{\bibinfo{person}{Jianan Li}, \bibinfo{person}{Xiaodan Liang}, \bibinfo{person}{Yunchao Wei}, \bibinfo{person}{Tingfa Xu}, \bibinfo{person}{Jiashi Feng}, {and} \bibinfo{person}{Shuicheng Yan}.} \bibinfo{year}{2017}\natexlab{}.
\newblock \showarticletitle{Perceptual generative adversarial networks for small object detection}. In \bibinfo{booktitle}{\emph{Proceedings of the IEEE conference on computer vision and pattern recognition}}. \bibinfo{pages}{1222--1230}.
\newblock


\bibitem[Li et~al\mbox{.}(2020)]%
        {li2020towards}
\bibfield{author}{\bibinfo{person}{Mengtian Li}, \bibinfo{person}{Yu-Xiong Wang}, {and} \bibinfo{person}{Deva Ramanan}.} \bibinfo{year}{2020}\natexlab{}.
\newblock \showarticletitle{Towards streaming perception}. In \bibinfo{booktitle}{\emph{European conference on computer vision}}. Springer, \bibinfo{pages}{473--488}.
\newblock


\bibitem[Lin et~al\mbox{.}(2025)]%
        {lin2025yolo}
\bibfield{author}{\bibinfo{person}{Ziyu Lin}, \bibinfo{person}{Yunfan Wu}, \bibinfo{person}{Yuhang Ma}, \bibinfo{person}{Junzhou Chen}, \bibinfo{person}{Ronghui Zhang}, \bibinfo{person}{Jiaming Wu}, \bibinfo{person}{Guodong Yin}, {and} \bibinfo{person}{Liang Lin}.} \bibinfo{year}{2025}\natexlab{}.
\newblock \showarticletitle{YOLO-LLTS: Real-Time Low-Light Traffic Sign Detection via Prior-Guided Enhancement and Multi-Branch Feature Interaction}.
\newblock \bibinfo{journal}{\emph{arXiv preprint arXiv:2503.13883}} (\bibinfo{year}{2025}).
\newblock


\bibitem[Liu et~al\mbox{.}(2020)]%
        {liu2020energy}
\bibfield{author}{\bibinfo{person}{Weitang Liu}, \bibinfo{person}{Xiaoyun Wang}, \bibinfo{person}{John Owens}, {and} \bibinfo{person}{Yixuan Li}.} \bibinfo{year}{2020}\natexlab{}.
\newblock \showarticletitle{Energy-based out-of-distribution detection}.
\newblock \bibinfo{journal}{\emph{Advances in neural information processing systems}}  \bibinfo{volume}{33} (\bibinfo{year}{2020}), \bibinfo{pages}{21464--21475}.
\newblock


\bibitem[Mishra et~al\mbox{.}(2025)]%
        {mishra2025benchmarking}
\bibfield{author}{\bibinfo{person}{Aditya Mishra}, \bibinfo{person}{Akshay Agarwal}, {and} \bibinfo{person}{Haroon Lone}.} \bibinfo{year}{2025}\natexlab{}.
\newblock \showarticletitle{Benchmarking Nighttime Traffic Sign Recognition with Illumination-Adaptive Detection and Semantic Attribute Reasoning}.
\newblock \bibinfo{journal}{\emph{arXiv preprint arXiv:2511.17183}} (\bibinfo{year}{2025}).
\newblock


\bibitem[{Mobileye}(2024)]%
        {mobileye_portfolio}
\bibfield{author}{\bibinfo{person}{{Mobileye}}.} \bibinfo{year}{2024}\natexlab{}.
\newblock \bibinfo{title}{Mobileye Product Portfolio}.
\newblock \bibinfo{howpublished}{\url{https://static.mobileye.com/website/corporate/docs/product_portfolio_2024.pdf}}.
\newblock
\newblock
\shownote{Accessed: Nov. 2026}.


\bibitem[{Mobileye}(2026)]%
        {mobileye_eyeq}
\bibfield{author}{\bibinfo{person}{{Mobileye}}.} \bibinfo{year}{2026}\natexlab{}.
\newblock \bibinfo{title}{{EyeQ} System-on-Chip Family}.
\newblock \bibinfo{howpublished}{\url{https://www.mobileye.com/technology/eyeq-chip/}}.
\newblock
\newblock
\shownote{Accessed: Nov. 2026}.


\bibitem[{NVIDIA}(2026)]%
        {nvidia_orinnano}
\bibfield{author}{\bibinfo{person}{{NVIDIA}}.} \bibinfo{year}{2026}\natexlab{}.
\newblock \bibinfo{title}{Jetson Orin Nano Super Developer Kit}.
\newblock \bibinfo{howpublished}{\url{https://marketplace.nvidia.com/en-us/enterprise/robotics-edge/jetson-orin-nano-super-developer-kit}}.
\newblock
\newblock
\shownote{Accessed: Nov. 2026}.


\bibitem[Redmon et~al\mbox{.}(2016)]%
        {redmon2016you}
\bibfield{author}{\bibinfo{person}{Joseph Redmon}, \bibinfo{person}{Santosh Divvala}, \bibinfo{person}{Ross Girshick}, {and} \bibinfo{person}{Ali Farhadi}.} \bibinfo{year}{2016}\natexlab{}.
\newblock \showarticletitle{You only look once: Unified, real-time object detection}. In \bibinfo{booktitle}{\emph{Proceedings of the IEEE conference on computer vision and pattern recognition}}. \bibinfo{pages}{779--788}.
\newblock


\bibitem[Ren et~al\mbox{.}(2015)]%
        {ren2015faster}
\bibfield{author}{\bibinfo{person}{Shaoqing Ren}, \bibinfo{person}{Kaiming He}, \bibinfo{person}{Ross Girshick}, {and} \bibinfo{person}{Jian Sun}.} \bibinfo{year}{2015}\natexlab{}.
\newblock \showarticletitle{Faster r-cnn: Towards real-time object detection with region proposal networks}.
\newblock \bibinfo{journal}{\emph{Advances in neural information processing systems}}  \bibinfo{volume}{28} (\bibinfo{year}{2015}).
\newblock


\bibitem[Sandler et~al\mbox{.}(2018)]%
        {sandler2018mobilenetv2}
\bibfield{author}{\bibinfo{person}{Mark Sandler}, \bibinfo{person}{Andrew Howard}, \bibinfo{person}{Menglong Zhu}, \bibinfo{person}{Andrey Zhmoginov}, {and} \bibinfo{person}{Liang-Chieh Chen}.} \bibinfo{year}{2018}\natexlab{}.
\newblock \showarticletitle{Mobilenetv2: Inverted residuals and linear bottlenecks}. In \bibinfo{booktitle}{\emph{Proceedings of the IEEE conference on computer vision and pattern recognition}}. \bibinfo{pages}{4510--4520}.
\newblock


\bibitem[Sermanet and LeCun(2011)]%
        {sermanet2011traffic}
\bibfield{author}{\bibinfo{person}{Pierre Sermanet} {and} \bibinfo{person}{Yann LeCun}.} \bibinfo{year}{2011}\natexlab{}.
\newblock \showarticletitle{Traffic sign recognition with multi-scale convolutional networks}. In \bibinfo{booktitle}{\emph{The 2011 international joint conference on neural networks}}. IEEE, \bibinfo{pages}{2809--2813}.
\newblock


\bibitem[Shankar et~al\mbox{.}(2021)]%
        {shankar2021image}
\bibfield{author}{\bibinfo{person}{Vaishaal Shankar}, \bibinfo{person}{Achal Dave}, \bibinfo{person}{Rebecca Roelofs}, \bibinfo{person}{Deva Ramanan}, \bibinfo{person}{Benjamin Recht}, {and} \bibinfo{person}{Ludwig Schmidt}.} \bibinfo{year}{2021}\natexlab{}.
\newblock \showarticletitle{Do image classifiers generalize across time?}. In \bibinfo{booktitle}{\emph{2021 IEEE/CVF International Conference on Computer Vision (ICCV)}}. IEEE, \bibinfo{pages}{9641--9649}.
\newblock


\bibitem[Stallkamp et~al\mbox{.}(2012)]%
        {stallkamp2012man}
\bibfield{author}{\bibinfo{person}{Johannes Stallkamp}, \bibinfo{person}{Marc Schlipsing}, \bibinfo{person}{Jan Salmen}, {and} \bibinfo{person}{Christian Igel}.} \bibinfo{year}{2012}\natexlab{}.
\newblock \showarticletitle{Man vs. computer: Benchmarking machine learning algorithms for traffic sign recognition}.
\newblock \bibinfo{journal}{\emph{Neural networks}}  \bibinfo{volume}{32} (\bibinfo{year}{2012}), \bibinfo{pages}{323--332}.
\newblock


\bibitem[Tabernik and Sko{\v{c}}aj(2019)]%
        {tabernik2019deep}
\bibfield{author}{\bibinfo{person}{Domen Tabernik} {and} \bibinfo{person}{Danijel Sko{\v{c}}aj}.} \bibinfo{year}{2019}\natexlab{}.
\newblock \showarticletitle{Deep learning for large-scale traffic-sign detection and recognition}.
\newblock \bibinfo{journal}{\emph{IEEE transactions on intelligent transportation systems}} \bibinfo{volume}{21}, \bibinfo{number}{4} (\bibinfo{year}{2019}), \bibinfo{pages}{1427--1440}.
\newblock


\bibitem[Tan and Le(2019)]%
        {tan2019efficientnet}
\bibfield{author}{\bibinfo{person}{Mingxing Tan} {and} \bibinfo{person}{Quoc Le}.} \bibinfo{year}{2019}\natexlab{}.
\newblock \showarticletitle{Efficientnet: Rethinking model scaling for convolutional neural networks}. In \bibinfo{booktitle}{\emph{International conference on machine learning}}. PMLR, \bibinfo{pages}{6105--6114}.
\newblock


\bibitem[Tran et~al\mbox{.}(2015)]%
        {tran2015learning}
\bibfield{author}{\bibinfo{person}{Du Tran}, \bibinfo{person}{Lubomir Bourdev}, \bibinfo{person}{Rob Fergus}, \bibinfo{person}{Lorenzo Torresani}, {and} \bibinfo{person}{Manohar Paluri}.} \bibinfo{year}{2015}\natexlab{}.
\newblock \showarticletitle{Learning spatiotemporal features with 3d convolutional networks}. In \bibinfo{booktitle}{\emph{Proceedings of the IEEE international conference on computer vision}}. \bibinfo{pages}{4489--4497}.
\newblock


\bibitem[Uikey et~al\mbox{.}(2024)]%
        {uikey2024indian}
\bibfield{author}{\bibinfo{person}{Rishabh Uikey}, \bibinfo{person}{Haroon~R Lone}, {and} \bibinfo{person}{Akshay Agarwal}.} \bibinfo{year}{2024}\natexlab{}.
\newblock \showarticletitle{Indian traffic sign detection and classification through a unified framework}.
\newblock \bibinfo{journal}{\emph{IEEE Transactions on Intelligent Transportation Systems}} \bibinfo{volume}{25}, \bibinfo{number}{10} (\bibinfo{year}{2024}), \bibinfo{pages}{14866--14875}.
\newblock


\bibitem[Uprety et~al\mbox{.}(2026)]%
        {uprety2026optimizing}
\bibfield{author}{\bibinfo{person}{Ishparsh Uprety}, \bibinfo{person}{Griffen Agnello}, {and} \bibinfo{person}{Xinghui Zhao}.} \bibinfo{year}{2026}\natexlab{}.
\newblock \showarticletitle{Optimizing deep learning based autonomous driving applications on edge devices}.
\newblock \bibinfo{journal}{\emph{Journal on Autonomous Transportation Systems}} \bibinfo{volume}{3}, \bibinfo{number}{3} (\bibinfo{year}{2026}), \bibinfo{pages}{1--18}.
\newblock


\bibitem[Wagner et~al\mbox{.}(2009)]%
        {wagner2009real}
\bibfield{author}{\bibinfo{person}{Daniel Wagner}, \bibinfo{person}{Gerhard Reitmayr}, \bibinfo{person}{Alessandro Mulloni}, \bibinfo{person}{Tom Drummond}, {and} \bibinfo{person}{Dieter Schmalstieg}.} \bibinfo{year}{2009}\natexlab{}.
\newblock \showarticletitle{Real-time detection and tracking for augmented reality on mobile phones}.
\newblock \bibinfo{journal}{\emph{IEEE transactions on visualization and computer graphics}} \bibinfo{volume}{16}, \bibinfo{number}{3} (\bibinfo{year}{2009}), \bibinfo{pages}{355--368}.
\newblock


\bibitem[Wang et~al\mbox{.}(2016)]%
        {wang2016temporal}
\bibfield{author}{\bibinfo{person}{Limin Wang}, \bibinfo{person}{Yuanjun Xiong}, \bibinfo{person}{Zhe Wang}, \bibinfo{person}{Yu Qiao}, \bibinfo{person}{Dahua Lin}, \bibinfo{person}{Xiaoou Tang}, {and} \bibinfo{person}{Luc Van~Gool}.} \bibinfo{year}{2016}\natexlab{}.
\newblock \showarticletitle{Temporal segment networks: Towards good practices for deep action recognition}. In \bibinfo{booktitle}{\emph{European conference on computer vision}}. Springer, \bibinfo{pages}{20--36}.
\newblock


\bibitem[Wojke et~al\mbox{.}(2017)]%
        {wojke2017simple}
\bibfield{author}{\bibinfo{person}{Nicolai Wojke}, \bibinfo{person}{Alex Bewley}, {and} \bibinfo{person}{Dietrich Paulus}.} \bibinfo{year}{2017}\natexlab{}.
\newblock \showarticletitle{Simple online and realtime tracking with a deep association metric}. In \bibinfo{booktitle}{\emph{2017 IEEE international conference on image processing (ICIP)}}. IEEE, \bibinfo{pages}{3645--3649}.
\newblock


\bibitem[Xu et~al\mbox{.}(2022)]%
        {xu2022mandheling}
\bibfield{author}{\bibinfo{person}{Daliang Xu}, \bibinfo{person}{Mengwei Xu}, \bibinfo{person}{Qipeng Wang}, \bibinfo{person}{Shangguang Wang}, \bibinfo{person}{Yun Ma}, \bibinfo{person}{Kang Huang}, \bibinfo{person}{Gang Huang}, \bibinfo{person}{Xin Jin}, {and} \bibinfo{person}{Xuanzhe Liu}.} \bibinfo{year}{2022}\natexlab{}.
\newblock \showarticletitle{Mandheling: Mixed-precision on-device dnn training with dsp offloading}. In \bibinfo{booktitle}{\emph{Proceedings of the 28th Annual International Conference on Mobile Computing And Networking}}. \bibinfo{pages}{214--227}.
\newblock


\bibitem[Zeng et~al\mbox{.}(2020)]%
        {zeng2020distream}
\bibfield{author}{\bibinfo{person}{Xiao Zeng}, \bibinfo{person}{Biyi Fang}, \bibinfo{person}{Haichen Shen}, {and} \bibinfo{person}{Mi Zhang}.} \bibinfo{year}{2020}\natexlab{}.
\newblock \showarticletitle{Distream: scaling live video analytics with workload-adaptive distributed edge intelligence}. In \bibinfo{booktitle}{\emph{Proceedings of the 18th Conference on Embedded Networked Sensor Systems}}. \bibinfo{pages}{409--421}.
\newblock


\bibitem[Zhang et~al\mbox{.}(2022a)]%
        {zhang2022bytetrack}
\bibfield{author}{\bibinfo{person}{Yifu Zhang}, \bibinfo{person}{Peize Sun}, \bibinfo{person}{Yi Jiang}, \bibinfo{person}{Dongdong Yu}, \bibinfo{person}{Fucheng Weng}, \bibinfo{person}{Zehuan Yuan}, \bibinfo{person}{Ping Luo}, \bibinfo{person}{Wenyu Liu}, {and} \bibinfo{person}{Xinggang Wang}.} \bibinfo{year}{2022}\natexlab{a}.
\newblock \showarticletitle{Bytetrack: Multi-object tracking by associating every detection box}. In \bibinfo{booktitle}{\emph{European conference on computer vision}}. Springer, \bibinfo{pages}{1--21}.
\newblock


\bibitem[Zhang et~al\mbox{.}(2022b)]%
        {zhang2022detection}
\bibfield{author}{\bibinfo{person}{Yanting Zhang}, \bibinfo{person}{Zijian Wang}, \bibinfo{person}{Ruoning Song}, \bibinfo{person}{Cairong Yan}, {and} \bibinfo{person}{Yonggang Qi}.} \bibinfo{year}{2022}\natexlab{b}.
\newblock \showarticletitle{Detection-by-tracking of traffic signs in videos}.
\newblock \bibinfo{journal}{\emph{Applied Intelligence}} \bibinfo{volume}{52}, \bibinfo{number}{7} (\bibinfo{year}{2022}), \bibinfo{pages}{8226--8242}.
\newblock


\bibitem[Zhu et~al\mbox{.}(2017)]%
        {zhu2017flow}
\bibfield{author}{\bibinfo{person}{Xizhou Zhu}, \bibinfo{person}{Yujie Wang}, \bibinfo{person}{Jifeng Dai}, \bibinfo{person}{Lu Yuan}, {and} \bibinfo{person}{Yichen Wei}.} \bibinfo{year}{2017}\natexlab{}.
\newblock \showarticletitle{Flow-guided feature aggregation for video object detection}. In \bibinfo{booktitle}{\emph{Proceedings of the IEEE international conference on computer vision}}. \bibinfo{pages}{408--417}.
\newblock


\bibitem[Zhu et~al\mbox{.}(2016)]%
        {zhu2016traffic}
\bibfield{author}{\bibinfo{person}{Zhe Zhu}, \bibinfo{person}{Dun Liang}, \bibinfo{person}{Songhai Zhang}, \bibinfo{person}{Xiaolei Huang}, \bibinfo{person}{Baoli Li}, {and} \bibinfo{person}{Shimin Hu}.} \bibinfo{year}{2016}\natexlab{}.
\newblock \showarticletitle{Traffic-sign detection and classification in the wild}. In \bibinfo{booktitle}{\emph{Proceedings of the IEEE conference on computer vision and pattern recognition}}. \bibinfo{pages}{2110--2118}.
\newblock


\end{thebibliography}

\clearpage
\section{Appendix}
\label{sec:appendix}

\subsection{Implementation Details}

This section provides complete implementation details for the Edge-TSR inference pipeline. The pipeline is implemented as a single Python process; the code for the full system and evaluation scripts is released alongside the video evaluation dataset. Table \ref{tab:hardware} gives the hardware and other related details of the Jetson Orin Nano used in the study. 

\subsubsection{Inference Pipeline}

The main inference loop is structured around a frame-level conditional: full detection and classification are invoked only on frames satisfying $t \bmod k = 0$; on all other frames, the previously computed bounding boxes and stabilized labels are propagated without any model invocation. This structure ensures that tracker state is updated at every frame but model inference is amortized over $k$ frames, achieving the throughput gains characterized in Section \ref{sec:results}.

\begin{table}[h]
\centering
\caption{NVIDIA Jetson Orin Nano hardware specifications.}
\label{tab:hardware}
\resizebox{0.95\columnwidth}{!}{
\begin{tabular}{ll}
\toprule
\textbf{Component} & \textbf{Specification} \\
\midrule
CPU & Arm Cortex-A78AE, 6-core @ 1.5 GHz \\
GPU & NVIDIA Ampere, 1024 CUDA cores, 32 Tensor Cores \\
Memory & 8 GB LPDDR5, unified CPU/GPU pool \\
Memory Bandwidth & 68 GB/s \\
AI Performance & 67 TOPS \\
Storage & 64 GB microSD \\
Power Mode & 10W (MAXN SUPER during deployment) \\
OS & Ubuntu 22.04 (JetPack 6.0) \\
Deep Learning Framework & PyTorch 2.10, CUDA 12.6 \\
\bottomrule
\end{tabular}
}
\end{table}

\noindent\textbf{Detection.} The YOLOv8 detector is loaded via the Ultralytics API and invoked using \texttt{yolo.track()} with \texttt{persist=True}, which maintains ByteTrack's internal state across calls. The detector is run at \texttt{imgsz=640} with a confidence threshold of 0.5. Detections with bounding box width or height below 15 pixels are discarded before crop extraction. The tracker output provides persistent integer track identities (\texttt{results.boxes.id}); frames where no tracks are assigned receive a sentinel value of $-1$. 

\noindent\textbf{RoI Pre-processing.} Each valid detection is padded 
by factor $p=0.20$ in both spatial dimensions before cropping:
\begin{equation*}
    dx = \lfloor (x_2 - x_1) \cdot p \rfloor, \quad
    x_1' = \max(0,\, x_1 - dx), \quad
    x_2' = \min(W,\, x_2 + dx)
\end{equation*}
and analogously for the vertical dimension. The padded crop is converted from BGR to RGB, resized to $224\times224$ using \texttt{T.Resize}, converted to a normalized tensor using ImageNet statistics ($\mu=[0.485, 0.456, 0.406]$, $\sigma=[0.229, 0.224, 0.225]$), and passed to the classifier as a single-sample batch. 

\noindent\textbf{Classification.} The ResNet-50 classifier is loaded with \texttt{weights=None} and its fully connected head replaced with a linear layer of output dimension equal to the number of training classes (41). Weights are loaded from a \texttt{.pth} checkpoint trained on Fold 0 of INTSD. The classifier runs in \texttt{torch.no\_grad()} context in FP32 precision. The predicted class and softmax confidence are extracted via \texttt{torch.max} on the softmax output. \\

\noindent\textbf{Temporal stabilization.} Track state is maintained in a Python dictionary keyed by integer track identity. Each entry holds a \texttt{deque(maxlen=5)} of \texttt{(label, confidence)} tuples, a locked label string (initialized to \texttt{None}), and the mean confidence of the locked label. The full hysteresis logic is described in Section \ref{sec:method}; the complete implementation is provided in the released code. 

\noindent\textbf{Geometric pruning.} After temporal stabilization, any detection assigned the \textit{advertisement} class with bounding box width or height below 50 pixels is reassigned to \textit{unknown}. This rule is applied after stabilization to prevent advertisement-class noise from entering the memory buffer.

\subsubsection{System Metric Logging}

System metrics are sampled at every frame and not only on detection frames. This provides a complete picture of resource consumption throughout the video. The following quantities are collected:

\begin{itemize}
    \item \textbf{FPS}: computed as a running mean ($\text{total frames} / \text{elapsed time}$) at each frame  and stored as a per-frame list for standard deviation computation.
    
    \item \textbf{Component latency}: wall-clock time in seconds for each detector invocation, each classifier invocation, and each draw call; stored as separate lists and converted to milliseconds at reporting time.
    
    \item \textbf{CPU utilization}: sampled via \texttt{psutil.cpu\_percent()} at each frame.
    
    \item \textbf{RAM utilization}: sampled via \texttt{psutil.virtual\_memory()\allowbreak.percent} at each frame.
    
    \item \textbf{CPU temperature}: read from \texttt{/sys/class/thermal/therm\allowbreak al\_zone0/temp} and divided by 1000 to convert from millidegrees to degrees Celsius; \texttt{None}  is returned if the file is unavailable.
    
    \item \textbf{GPU utilization and temperature}: sampled via \texttt{jtop}, a dedicated Jetson hardware monitoring library that interfaces with the NVIDIA Tegra system-on-chip telemetry layer. A separate monitoring process runs concurrently with the inference pipeline, polling \texttt{jetson.stats} at 1-second intervals. GPU utilisation is read from the \texttt{``GPU''} field and GPU temperature from \texttt{``Temp gpu''}; both are stored as per-sample lists. Mean and maximum values are computed post-hoc over all non-null samples. This approach is necessary because the Jetson Orin Nano's GPU metrics are not accessible via \texttt{psutil}, which only exposes CPU-side statistics on this platform.

    \item \textbf{GPU memory utilization}: read from the \texttt{``RAM''} field of \texttt{jetson.stats}, which on the Jetson Orin Nano reflects the unified LPDDR5 memory pool shared between CPU and GPU. When the field returns a dictionary, used and total memory are extracted separately; when it returns a scalar, it is interpreted as a percentage utilization directly. Mean and maximum utilization are reported as percentages of the 8 GB unified pool.
\end{itemize}

All lists are aggregated post-hoc using \texttt{numpy} (\texttt{np.mean}, \texttt{np.std}, \texttt{np.max}) and written to a JSON results file alongside ablation metadata.

\subsubsection{Output Format}

The pipeline writes four output files per run, all in JSON format:

\begin{itemize}
    \item \textbf{\texttt{predictions.json}}: a list of per-frame objects, each containing \texttt{frame\_id} and a list of \texttt{detections}. Each detection record contains \texttt{bbox} ($[x_1, y_1, x_2, y_2]$ in pixel coordinates), \texttt{det\_conf} (detector confidence), \texttt{cls\_conf} (classifier softmax confidence), \texttt{score} (product of  both), \texttt{cls\_label} (stabilized class string), and \texttt{track\_id} (integer ByteTrack identity).

    \item \textbf{\texttt{evaluated\_frames.json}}: a list of integer frame indices on which full detection and classification were invoked (i.e., frames satisfying $t \bmod k = 0$). This file is required by the evaluation scripts to distinguish inferred frames from propagated frames when computing per-frame metrics.

    \item \textbf{\texttt{inference\_results.json}}: aggregate system metrics for the run, including throughput, latency breakdown, CPU/GPU statistics, and ablation condition flags. The schema is identical across all ablation variants to enable direct comparison.

    \item \textbf{\texttt{gpu\_stats.json}}: This file aggregates GPU and unified memory statistics collected by a concurrent monitoring process during inference. It reports mean GPU utilization, mean and maximum GPU temperature ($\degree$ C), and mean and maximum utilization of the unified LPDDR5 memory as a percentage of total capacity. Metrics are computed over all valid samples, with the \texttt{samples} field indicating full monitoring coverage. The file is generated independently and later cross-referenced with \texttt{inference\_results.json} to produce the system-level results in Tables~\ref{tab:system_performance}.
\end{itemize}

\subsection{Latency Decomposition per Sampling Interval}
\label{sec:appendix_latency}

\textbf{Setup.} All measurements use the Dense urban video (27{,}000 frames after a 30-frame warm-up), processed offline and headless on the Jetson Orin Nano in MAXN SUPER mode with FP32 PyTorch models, one run per $k$. Each stage timer is stopped only after CUDA synchronization. Following the definitions in Section~\ref{sec:system_evaluation}, a sampled frame executes Det+Trk, classification of each detected region, and stabilization; a non-sampled frame only propagates the most recent boxes and stabilized labels. $L_s$ and $L_p$ are the mean end-to-end latencies of sampled and non-sampled frames, including frame decoding and overlay rendering, and $L_{\text{avg}} = (L_s + (k-1)L_p)/k$.

\begin{table*}[t]
\centering
\caption{Latency decomposition per sampling interval $k$ (Dense urban video, offline, headless, FP32). Det+Trk, Cls (per region), and Stab are mean per-invocation latencies; $\bar{n}$ is the mean number of classified regions per sampled frame; $L_s$, $L_p$, and $L_{\text{avg}}$ are mean per-frame latencies. All latencies in ms. $\sigma$ is the standard deviation of FPS over 1-s windows, and CV $= \sigma/\text{FPS}$.}
\label{tab:latency_k}
\resizebox{0.8\textwidth}{!}{
\begin{tabular}{lccccccccccc}
\toprule
$\textbf{k}$ & \textbf{Det+Trk} & \textbf{Cls} & \textbf{Stab} & $\bar{\textbf{n}}$ & \textbf{$L_s$ }& \textbf{$L_p$ }& \textbf{$L_{\text{avg}}$} & $\textbf{1000/\text{FPS}}$ & \textbf{FPS} & \textbf{$\sigma$} & \textbf{CV} \\
\midrule
1  & 30.22 & 20.98 & 0.07 & 1.92 & 75.42 & --   & 75.42 & 75.43 & 13.26  & 7.03  & 0.53 \\
2  & 33.80 & 21.43 & 0.08 & 1.88 & 79.59 & 4.84 & 42.22 & 42.22 & 23.68  & 10.27 & 0.43 \\
3  & 36.77 & 21.85 & 0.07 & 1.83 & 83.18 & 5.82 & 31.61 & 31.61 & 31.63  & 11.31 & 0.36 \\
4  & 38.65 & 22.55 & 0.07 & 1.76 & 85.21 & 5.62 & 25.52 & 25.53 & 39.18  & 12.26 & 0.31 \\
5  & 40.59 & 23.25 & 0.07 & 1.72 & 87.72 & 6.07 & 22.40 & 22.41 & 44.62  & 12.50 & 0.28 \\
10 & 50.02 & 27.78 & 0.07 & 1.51 & 99.19 & 6.68 & 15.93 & 15.94 & 62.73  & 12.64 & 0.20 \\
30 & 53.91 & 28.77 & 0.06 & 1.35 & 99.30 & 5.91 & 9.02  & 9.03  & 110.72 & 15.07 & 0.14 \\
\bottomrule
\end{tabular}}
\end{table*}

\begin{table}[t]
    \centering
    \caption{Linear fit of sampled-frame compute (Det+Trk + Cls + Stab) against the number of classified regions $n$ per frame: compute $\approx a + b\,n$.}
    \label{tab:load_fit}
    \begin{tabular}{lccc}
    \toprule
    $\textbf{k}$ & $\textbf{a}$ \textbf{(ms)} & $\textbf{b}$ \textbf{(ms/region)} & \textbf{Pearson} $\textbf{r}$ \\
    \midrule
    1  & 30.77 & 20.74 & 0.996 \\
    2  & 36.61 & 19.98 & 0.995 \\
    3  & 41.05 & 19.55 & 0.991 \\
    4  & 44.04 & 19.53 & 0.983 \\
    5  & 46.45 & 19.88 & 0.980 \\
    10 & 59.57 & 21.52 & 0.947 \\
    30 & 55.21 & 27.87 & 0.966 \\
    \bottomrule
    \end{tabular}
    
\end{table}

\textbf{Reconciling latency and throughput.} For every $k$, the amortized per-frame latency $L_{\text{avg}}$ agrees with the inverse of the independently measured wall-clock throughput to within 0.01~ms (Table~\ref{tab:latency_k}), so the reported FPS is fully explained by the cost of sampled and non-sampled frames. Throughput gains from sparse sampling therefore come entirely from replacing sampled frames ($L_s = 75$--$99$~ms) with propagation-only frames ($L_p = 4.8$--$6.7$~ms).

\textbf{Propagation-only cost.} A non-sampled frame costs $L_p = 4.8$--$6.7$~ms, of which frame decoding accounts for 3.3--5.0~ms and overlay rendering for 0.2~ms; no detector, tracker, or classifier is invoked. Temporal stabilization adds only 0.06--0.08~ms per sampled frame, less than 0.1\% of $L_s$.

\textbf{Load-proportional cost.} Sampled-frame compute grows linearly with the number of classified regions (Table~\ref{tab:load_fit}). For $k \leq 5$, each region adds 19.5--20.7~ms with $r \geq 0.98$, closely matching the per-invocation classifier latency, since regions are classified sequentially. At $k=1$, a frame without detections costs $\approx$31~ms of compute, whereas a frame with the observed maximum of 11 regions costs $\approx$259~ms. This dependence on scene content, rather than on the sampling interval, explains why the standard deviation of $L_s$ remains at 36--43~ms for all $k$, and why throughput differs across driving conditions (Section~\ref{sec:results}).

\textbf{Per-invocation latency and region count versus $k$.} Two secondary trends are visible in Table~\ref{tab:latency_k}. First, per-invocation Det+Trk and Cls latencies increase with $k$ (30.2 to 53.9~ms and 21.0 to 28.8~ms), which is consistent with the GPU operating at lower clock frequencies during the longer idle intervals between sparse invocations. Second, the number of classified regions per sampled frame decreases with $k$ (1.92 to 1.35), as larger inter-frame motion causes more tracker association failures, so fewer tracks are confirmed and output. This reduction also contributes to the accuracy decline at large $k$ (Figure~\ref{fig:sampling_tradeoff}).

\subsection{Reduced-Precision Deployment with TensorRT}
\label{sec:appendix_trt}

\textbf{Motivation and setup.} To test whether the throughput limits reported in Section~\ref{sec:results} are an artifact of unoptimized FP32 PyTorch execution, we deploy both the detector and the classifier as TensorRT engines in FP32 and FP16 precision [engine export procedure, e.g., detector exported via Ultralytics, classifier via ONNX and \texttt{trtexec}]. All other pipeline components (tracking, temporal stabilization, and pre- and post-processing) are unchanged, and all variants run at $k=3$ on the same video under the offline protocol of Section~\ref{sec:system_evaluation}.

\textbf{Results.} TensorRT increases offline throughput by only 9.8\% (31.63 $\rightarrow$ 34.72~FPS), and FP16 provides no measurable gain over TensorRT FP32 (34.72 vs.\ 34.65~FPS; Table~\ref{tab:quantization}). Per-region classification latency decreases from 21.85 to 15.67~ms ($-28\%$), whereas Det+Trk latency increases from 34.96 to 46.74~ms ($+34\%$), so the gains of the classifier are partially offset at the pipeline level. Recognition quality is largely preserved: mAP@50 decreases by 0.16~pp, mAP@50:95 by at most 0.27~pp, and classification accuracy by at most 0.24~pp, while A.P. and A.R. decrease by approximately 3~pp and 1.6--1.9~pp, respectively. CPU utilization increases under TensorRT (39.64 $\rightarrow$ 46.84--47.56\%), whereas CPU temperatures are lower; the latter may partly reflect run-to-run differences in ambient conditions.

\textbf{Inference.} Three observations indicate that end-to-end latency is dominated by host-side processing rather than network computation. First, accelerating network inference by switching to TensorRT and halving the numerical precision yields less than 10\% end-to-end gain, and FP16 yields none over FP32. Second, even with TensorRT, classifying a single region still costs $\approx$16~ms, which is consistent with the CPU-bound crop preprocessing (color conversion, resizing, and tensor conversion) executed per region before inference. Third, CPU utilization rises as GPU inference becomes faster, suggesting that the host becomes the bottleneck. The increase in Det+Trk latency is likewise not attributable to network computation; we attribute it to [cause if known, e.g., the overhead of the engine wrapper and host-device transfers], which we did not isolate further. Since TensorRT offers only a marginal throughput benefit at a cost of about 3~pp in classification precision, we report all main results with the PyTorch FP32 models, which keeps the pipeline simpler and fully reproducible. Moving crop preprocessing to the GPU and batching region classification are more promising optimizations than reduced precision for this pipeline.

\begin{table}[t]
\centering
\caption{Effect of TensorRT deployment on Edge-TSR at sampling interval $k=3$ (offline, video test set, 27{,}030 frames, Jetson Orin Nano). Det+Trk and Cls are per-invocation latencies; Cls is per region.}
\label{tab:quantization}
\begin{tabular}{lccc}
\toprule
 & \textbf{PyTorch FP32} & \textbf{TRT FP32} & \textbf{TRT FP16} \\
\midrule
\multicolumn{4}{l}{\textit{Throughput and latency}} \\
FPS$_{\text{off}}$ $\uparrow$        & 31.63 & 34.65 & \textbf{34.72} \\
Det+Trk (ms) $\downarrow$            & \textbf{34.96} & 46.89 & 46.74 \\
Cls (ms) $\downarrow$                & 21.85 & 15.73 & \textbf{15.67} \\
\midrule
\multicolumn{4}{l}{\textit{Resources}} \\
CPU util. (\%)                       & \textbf{39.64} & 46.84 & 47.56 \\
Avg. CPU Temp. (\textdegree C)  & 54.28 & \textbf{48.11}  & 48.60 \\
Max. CPU Temp. (\textdegree C)  & 57.96 & \textbf{49.06} & 49.75 \\
\midrule
\multicolumn{4}{l}{\textit{Performance}} \\
mAP@50 $\uparrow$                    & \textbf{73.12} & 72.96 & 72.96 \\
mAP@50:95 $\uparrow$                 & \textbf{45.79} &  45.61 & 45.52 \\
Cls. accuracy $\uparrow$             & \textbf{75.33} & 75.09 & 75.11 \\
A.P. $\uparrow$                  & \textbf{75.23} & 72.18 & 72.31 \\
A.R. $\uparrow$                  & \textbf{72.10} & 70.53 & 70.22 \\
\bottomrule
\end{tabular}
\end{table}

\subsection{Adaptive Sampling Interval}
\label{sec:appendix_adaptive}

\textbf{Motivation.} Edge-TSR uses a fixed sampling interval $k=3$. A natural alternative, in the spirit of adaptive video analytics systems that tune configurations at runtime~\cite{jiang2018chameleon}, is to adapt $k$ to the measured load. We evaluate two such controllers.

\textbf{Load-adaptive controller.} The controller maintains exponential moving averages of the measured cost of sampled and non-sampled frames and uses the latency model of Appendix~\ref{sec:appendix_latency} to predict the amortized per-frame cost $(L_s + (k-1)L_p)/k$. With a real-time budget equal to the camera rate (30~FPS, 33.3~ms per frame), $k$ is increased by one when the predicted cost exceeds the budget, and decreased by one when $k-1$ fits within 80\% of the budget. Changes are separated by a cooldown of five sampled frames, $k$ is bounded to $[3, 7]$, and a thermal guard increases $k$ above 80\textdegree C (never triggered in our runs).

\textbf{Accuracy-constrained controller.} This variant additionally bounds $k$ from above using two scene-level signals. First, a lifetime bound $k \leq L/3$, where $L$ is the 20th percentile of recent track lifetimes in frames, ensures that briefly visible signs still receive the three observations required for label locking. Second, an under-sampled track rate, i.e., the fraction of terminated tracks observed fewer than three times, triggers a back-off of $k$ by two when it exceeds 20\% and blocks increases when it lies between 10\% and 20\%. When the load and accuracy constraints are compatible, the controller selects the largest admissible $k$; when they conflict, real-time operation takes priority.

\textbf{Setup.} Both controllers start at $k=3$ and are evaluated on the Dense urban video under the same protocol as Table~\ref{table:results}.

\begin{table*}[t]
    \centering
    \footnotesize
    \caption{Effect of making the sampling rate ``k'' adaptive with respect to load. The constraint means is put on to avoid increasing ``k'' too much which would adveresly affect the detection and classification performance.}
    \label{tab:adaptive_k}
    \setlength{\tabcolsep}{6pt}
    \resizebox{0.9\textwidth}{!}{
    \begin{tabular}{p{4cm}ccccccc}
    \toprule
    \textbf{Model} & \multicolumn{2}{c}{\textbf{Detection (\%)}} & \multicolumn{3}{c}{\textbf{Classification (\%)}} & \textbf{FPS} & $\bar{k}$ \\
    \cmidrule(lr){2-3} \cmidrule(lr){4-6} 
    & \textbf{mAP@50} & \textbf{mAP@50:95} & \textbf{A.P.} & \textbf{A.R.} & \textbf{Acc.} & \textbf{(end-to-end)} \\
    \midrule
    
    \multicolumn{7}{c}{\textit{Video-based Evaluation}} \\
    \midrule
    Edge-TSR (k = 3) & \textbf{73.12} & 45.79 & 75.23 & \textbf{72.10} & \textbf{75.33} & \textbf{31.63} & 3.00 \\
    Adaptive Edge-TSR & 65.62 & 37.35 & \textbf{76.52} & 52.85 & 59.31 & 35.05 & 4.33\\
    Adaptive Edge-TSR w/ constraint & 62.32 & 35.52 & 72.15 & 46.54 & 57.22 & 35.88 & 4.76 \\
    \bottomrule
    \end{tabular}
    }
\end{table*}

\textbf{Results.} Both controllers trade a small throughput gain for a large loss in recognition quality (Table~\ref{tab:adaptive_k}). The load-adaptive controller increases throughput by 10.8\% (31.63 $\rightarrow$ 35.05~FPS) but reduces recall by 19.25~pp and accuracy by 16.02~pp, with a slight gain in precision (+1.29~pp). The accuracy-constrained controller increases throughput by 13.4\% but reduces recall by 25.56~pp and accuracy by 18.11~pp.

\textbf{Inference.} Three factors explain these results. First, since $k=3$ already maximizes F1 (Figure~\ref{fig:sampling_tradeoff}(b)) and the controllers never select $k<3$, any adaptation moves the system into the regime where accuracy declines. Second, because per-frame cost grows by $\approx$20~ms per classified region (Table~\ref{tab:load_fit}), load-driven increases of $k$ occur in sign-dense segments, precisely where skipped observations most reduce recall. Third, the accuracy signals are themselves degraded by sparse sampling: at larger $k$, tracks fragment and are confirmed less often, which biases measured lifetimes and under-sampling rates. Combined with selecting the largest admissible $k$, this makes the constrained controller more aggressive rather than more conservative. For continuous TSR, a fixed interval at the accuracy optimum is therefore preferable to load-adaptive sampling; adaptive control would be more useful on platforms where the optimal interval is not throughput-feasible, in which case $k_{\min}$ from Section~\ref{sec:discussion} provides the starting point.

\subsection{Backbone Comparison}

Table~\ref{tab:backbone} reports the performance of three candidate classification backbones evaluated on the dense urban traffic scenario under identical streaming video conditions on the Jetson Orin Nano. All backbones are trained on Fold 0 of INTSD using the same hyperparameters, crop preprocessing pipeline ($p=0.20$ padding, $224\times224$ resize, ImageNet normalization). Latency is measured as mean per-crop wall-clock inference time on the Jetson GPU in FP32 precision, averaged over all classification invocations in the evaluation video.

\begin{table}[htbp]
\centering
\caption[Backbone comparison for Edge-TSR]{Backbone comparison for fine-grained classification. All models evaluated under identical conditions on the Jetson Orin Nano.}
\label{tab:backbone}
\resizebox{\columnwidth}{!}{
\begin{tabular}{lcccccc}
\toprule
\textbf{Backbone} 
& \textbf{A.P. (\%)} 
& \textbf{A.R. (\%)} 
& \textbf{Acc. (\%)}
& \textbf{FPS}
& \textbf{Det + Trk (ms)} 
& \textbf{Cls (ms)} \\
\midrule
MobileNetV3 \cite{howard2019searching}  & 52.42 & 73.36 & 72.68 & 24.52 & 40.09 & 21.69 \\
EfficientNet-B3 \cite{tan2019efficientnet} & 61.53 & \textbf{75.67} & 73.08 & 19.26 & 34.75 & 41.80 \\
ResNet-50 \cite{he2016deep} & \textbf{75.23} & 72.10 & \textbf{75.33} & \textbf{27.01} & 34.96 & 21.85 \\
\bottomrule
\end{tabular}
}
\end{table}

The results reveal two findings that directly justify the ResNet-50 choice. First, the A.P. gap is large and asymmetric: ResNet-50 outperforms MobileNetV3 by 22.81\% and EfficientNet-B3 by 13.70\%, while the accuracy gap is smaller but consistent (2.65\% and 2.25\% respectively). The recall figures partially mask this: EfficientNet-B3 achieves the highest A.R. (75.67\%), marginally above ResNet-50 (72.10\%), but this recall advantage does not translate into classification accuracy, indicating that EfficientNet-B3 produces more detections at the cost of precision — a trade-off that is unfavorable in a fine-grained recognition setting where label correctness is the primary requirement. 

Second, and more counterintuitively, ResNet-50 achieves the \textit{highest} end-to-end FPS (27.01) of the three backbones, despite having the largest parameter count. EfficientNet-B3 is the slowest at 19.26 FPS, with per-crop classifier latency of 41.80 ms — nearly double that of ResNet-50 (21.85 ms) and MobileNetV3 (21.69 ms). This result reflects a known property of EfficientNet on embedded GPU hardware: its compound scaling strategy, which balances depth, width, and resolution, introduces memory access patterns and activation shapes that are less efficiently handled by the Jetson Ampere architecture than the straightforward residual blocks of ResNet-50. MobileNetV3's per-crop latency (21.69 ms) is essentially identical to ResNet-50 (21.85 ms) on this hardware, meaning its lightweight design yields no practical speed benefit on the Jetson while incurring a 22.81\% A.P. penalty. We therefore adopt ResNet-50 as the classification backbone: it is simultaneously the most accurate, the fastest end-to-end, and the most stable under the fine-grained discrimination demands of the 41-class Indian traffic sign vocabulary.

\subsection{Hyperparameter Sensitivity}

The temporal stabilization module uses fixed hyperparameters ($T=5$, locking threshold $\geq 3$ observations, confidence threshold 60\%, escape margin $\Delta = 5\%$) selected empirically on the dense urban scenario. These parameters may not be optimal across all operating conditions — in particular, high-speed driving may warrant smaller $T$ and lower locking thresholds to avoid temporal window starvation, while low-speed urban crawl may benefit from larger windows. An adaptive parameter schedule conditioned on vehicle speed or detected scene complexity is a natural extension. Please refer to \ref{tab:hyperparams_edgetsr} for the full list of the hyperparameters along with their optimum value.

\begin{table}[t]
\centering
\caption[Hyperparameters of Edge-TSR]{Edge-TSR inference pipeline hyperparameters.}
\label{tab:hyperparams_edgetsr}
\resizebox{0.95\columnwidth}{!}{
\begin{tabular}{ll}
\toprule
\textbf{Parameter} & \textbf{Value} \\
\midrule
Detector             & YOLOv8 (custom weights, \texttt{yolov8s} backbone) \\
Detector input size  & $640 \times 640$ \\
Detector conf. threshold & 0.5 \\
Tracker              & ByteTrack (\texttt{bytetrack.yaml}, default params) \\
Sampling interval ($k$) & 3 \\
Min. box size ($d_{min}$)        & $15 \times 15$ px \\
Classifier           & ResNet-50, FP32, 41 classes \\
Classifier input     & $224 \times 224$, ImageNet normalization \\
$d_{adv}$ & 50 \\
Padding factor ($p$)   & 0.20 \\
Memory buffer ($T$)    & 5 frames \\
Lock threshold       &  $\tau = 3$, $\delta = 0.6$ \\
Escape margin ($\Delta$) & 0.05 \\
Adv. pruning threshold & width or height $< 50$ px \\
\bottomrule
\end{tabular}
}
\end{table}

\subsection{Qualitative Results}
\label{sec:appen_qualitative_res}

\subsubsection{Dense}
\begin{figure*}[t]
    \centering
    \includegraphics[width=0.9\textwidth]{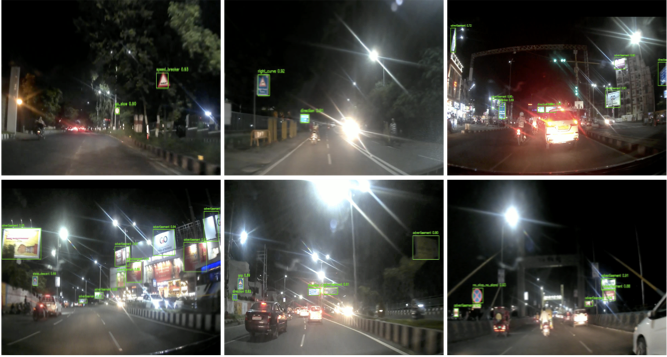}
    \caption{Qualitative results of Edge-TSR on the dense urban traffic scenario. The system correctly detects and classifies multiple sign categories simultaneously under nighttime conditions with high scene clutter, oncoming headlight glare, and co-occurring advertisement boards. High detector confidence scores are sustained even in frames containing up to 11 simultaneous detections.}
    \label{fig:dense}
\end{figure*}

Figure~\ref{fig:dense} demonstrates Edge-TSR's detection and classification performance under the dense urban traffic condition — the most challenging scenario in terms of simultaneous object count, background clutter, and illumination variability. The system successfully localizes and correctly classifies multiple traffic sign categories within a single frame, including regulatory signs (\textit{no stop no stand}, \textit{go slow}), warning signs (\textit{speed breaker}, \textit{right curve}), and directional signs (\textit{direction}), even in the presence of a high density of visually confusable \textit{advertisement} boards that trigger parallel detector responses.

Confidence scores for correctly classified traffic signs remain high ($80--95\%$) across frames with up to 11 simultaneous detections, demonstrating that the decoupled detection-classification architecture maintains fine-grained discrimination quality under maximum classification invocation load.

\subsubsection{Rain}
\begin{figure*}[t]
    \centering
    \includegraphics[width=0.9\textwidth]{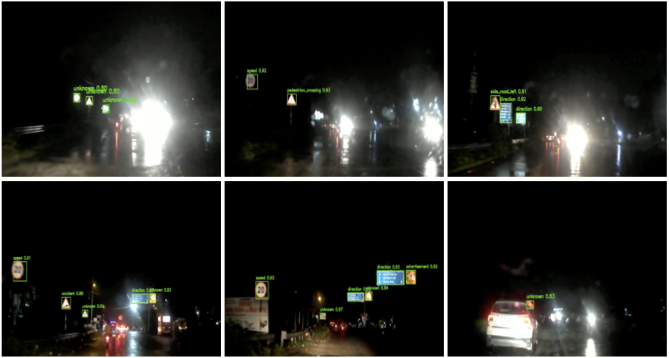}
    \caption{Qualitative results of Edge-TSR on the rain scenario. The system correctly detects and classifies signs under active precipitation, despite severe lens occlusion from water droplets, specular highlights from wet road surfaces, and oncoming headlight glare that dominates several frames. The bottom-right frame illustrates a characteristic failure mode under heavy rain: a vehicle's rear brake light reflected in water droplets on the lens is incorrectly triggered as a sign detection (labeled \textit{unknown}, confidence 83\%), demonstrating the limits of the class-agnostic detector under extreme precipitation artifacts.}
    \label{fig:rain}
\end{figure*}

Figure~\ref{fig:rain} shows Edge-TSR's behavior under the rain scenario — the condition that most severely degrades detector confidence through concurrent lens-level and scene-level degradations. Despite active precipitation introducing water droplet occlusion directly on the camera lens, specular highlights from wet road surfaces, and contrast reduction from reflected headlight glare, the system successfully localizes and correctly classifies a range of sign categories including \textit{pedestrian crossing}, \textit{side road left}, \textit{direction}, \textit{accident}, and speed limit signs, with confidence scores sustained between 85\% and 93\% on correctly classified instances. The \textit{unknown}-class outputs visible in several frames (top-left, bottom-center, bottom-right) reflect the hysteresis module's correct behavior under uncertainty: detections that do not accumulate sufficient consistent evidence within the temporal window $T{=}5$ are withheld as \textit{unknown} rather than committed to a potentially incorrect label, demonstrating that the stabilization module degrades gracefully under confidence suppression rather than producing spurious high-confidence misclassifications.

The bottom-right frame illustrates the primary failure mode under heavy rain: a vehicle's rear brake light, diffused and bloomed through water droplets adhering to the camera lens, generates a detector response that is assigned the \textit{unknown} label at confidence 83\%.

This false positive is mechanistically explicable: the bloomed brake light presents as a circular, red-dominant region of high intensity — a visual signature that closely resembles the appearance of a \textit{speed} sign (circular red background) or a \textit{stop} sign (octagon red background) at the detector's $640 \times 640$ input resolution, causing the class-agnostic detector to fire. Crucially, the classifier correctly withholds a committed label, assigning \textit{unknown} rather than erroneously locking onto \textit{speed} or \textit{stop} — demonstrating that the closed-world classifier retains sufficient discriminative capacity to reject a visually plausible but contextually incorrect match. This failure mode is therefore partially mitigated by the classification stage, but the false detection itself is irreducible at the detector level: no amount of temporal stabilization can suppress a detector response that recurs consistently across frames, as repeated \textit{unknown} observations would accumulate in the track buffer and consume window capacity without contributing label information. Mitigation would require either a rain-aware detection backbone that suppresses high-intensity bloomed regions, or a lens contamination detector that reduces detector sensitivity during severe precipitation events.

\subsubsection{Rural}
\begin{figure*}[t]
    \centering
    \includegraphics[width=0.9\textwidth]{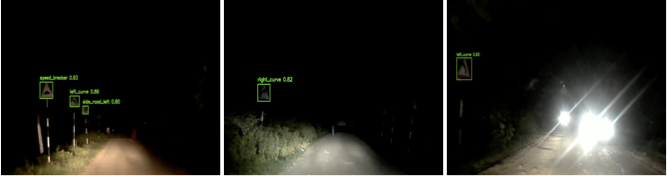}
    \caption{Qualitative results of Edge-TSR on the rural/dark scenario, where illumination is provided exclusively by the vehicle's headlights with no street lighting. The system correctly detects and classifies warning signs against near-zero background luminance, relying entirely on retroreflective sign surfaces illuminated by headlights. The rightmost frame demonstrates correct detection and classification under simultaneous oncoming headlight glare where high dynamic range between the bright glare source and the dark sign background compresses the effective contrast available to the detector.}
    \label{fig:rural}
\end{figure*}

Figure~\ref{fig:rural} presents qualitative results under the rural/dark condition — the most extreme illumination scenario in the evaluation set, where no street lighting is present and sign visibility depends entirely on retroreflective surfaces illuminated by the vehicle's own headlights. Despite near-zero background luminance and the absence of any ambient light source, Edge-TSR successfully localizes and correctly classifies multiple warning sign categories across all three frames, including \textit{speed breaker}, \textit{left curve}, \textit{side road left}, and \textit{right curve}, with confidence scores ranging from 80\% to 93\%.

\subsubsection{OOD}
\begin{figure*}[t]
    \centering
    \includegraphics[width=0.9\textwidth]{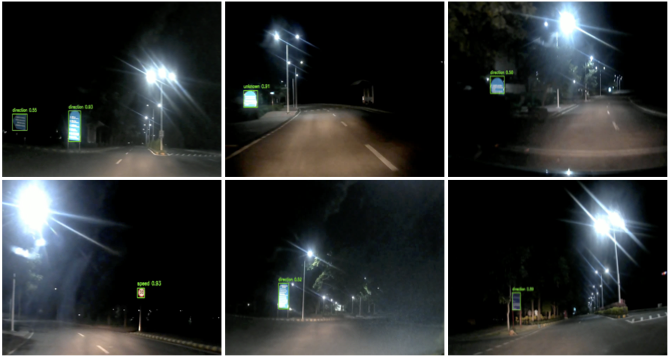}
    \caption{Qualitative results of Edge-TSR on the out-of-distribution (OOD) scenario, containing sign categories rendered in non-standard physical forms and sign categories absent from the 41-class training vocabulary. The system correctly detects and classifies in-vocabulary signs including \textit{direction} and \textit{speed} where sufficient visual similarity to training examples exists, and correctly withholds commitment on a genuinely out-of-vocabulary green directional board (\textit{unknown}, 91\%, top-center). The bottom-right frame illustrates a characteristic high-confidence misclassification failure: a large decorative blue roadside board bearing text is incorrectly classified as \textit{direction} at confidence 89\%, representing the most dangerous OOD failure mode — a spuriously high-confidence committed label on a non-sign object whose blue rectangular form and white text layout closely mirror the visual signature of the \textit{direction} training class.}
    \label{fig:ood}
\end{figure*}

Two properties of this result are worth noting in the context of the system-level findings reported in Section~\ref{sec:results}. First, the successfully detected signs in the left and center frames are small, dark, and low-contrast against the surrounding vegetation — the detector must identify retroreflective triangle outlines against a background that is nearly identical in luminance to the sign interior, operating at the lower bound of what the $640 \times 640$ input resolution can resolve.

The high confidence scores on these detections indicate that the INTSD-trained detector has internalized retroreflective appearance cues specific to nighttime sign illumination, rather than relying on ambient contrast that is absent in this condition. Second, the rightmost frame introduces simultaneous oncoming headlight glare — a high dynamic range condition where the bright glare source and the dark sign background differ by several orders of magnitude in luminance. Edge-TSR correctly detects and classifies \textit{left curve} at confidence 92\% in this frame, demonstrating that the pipeline remains functional even when the dominant light source in the scene is an oncoming vehicle rather than the host vehicle's own headlights. This result directly supports the finding in Section~\ref{sec:results} that the rural/dark bottleneck is localized to the \textit{detection} stage rather than the classification stage: when the detector successfully localizes a sign under these conditions, the classifier operating on the normalized crop reliably assigns the correct label regardless of absolute scene illumination level.

Figure~\ref{fig:ood} presents qualitative results under the OOD scenario, which encompasses two distinct generalization challenges as characterized in Section~\ref{sec:results}: intra-class distribution shift, where in-vocabulary sign categories appear in non-standard physical forms, and true open-set failure, where the encountered sign or board has no valid equivalent in the 41-class training vocabulary.

Both failure modes are visible across the six frames. The top-center frame shows the system's correct handling of a genuinely out-of-vocabulary sign: a large blue directional board of a format not present in INTSD is detected but assigned \textit{unknown} at confidence 91\%, indicating that the classifier correctly rejects the nearest in-vocabulary match rather than committing to a plausible but incorrect label. 

Similarly, the bottom-center frame shows a large blue directional board assigned \textit{direction} at confidence 52\% — a low-confidence committed label that reflects the classifier's uncertainty when faced with a sign whose layout and typography differ substantially from training examples, and which would likely be overridden by the hysteresis escape condition under sustained observation. The top-right frame shows a further instance of this pattern: an OOD directional sign assigned \textit{direction} at confidence 50\%.

The bottom-right frame illustrates the most consequential OOD failure mode: a large decorative blue roadside board bearing text (physically resembling a direction sign in color, size, and shape) is classified as \textit{direction} at high confidence (89\%). This is a spuriously high-confidence misclassification: the classifier assigns a committed label well above the locking threshold ($\delta = 0.60$), meaning the temporal stabilization module would lock this incorrect label and maintain it across subsequent frames.

The failure is mechanistically distinct from the low-confidence OOD cases above: the board shares sufficient visual features with the \textit{direction} training distribution — blue rectangular background, white text, similar aspect ratio — that the classifier's softmax output is dominated by a single class rather than being distributed across alternatives. This is precisely the failure mode identified in Section~\ref{sec:results} as unreachable by the geometric pruning rule or the unknown-class suppression logic: both mechanisms operate on low-confidence or small detections, and neither can intercept a large, high-confidence, visually plausible misclassification.

\end{document}